\def\year{2021}\relax
\documentclass[letterpaper]{article} 
\usepackage{aaai21}  
\usepackage{times}  
\usepackage{helvet} 
\usepackage{courier}  
\usepackage[hyphens]{url}  
\usepackage{graphicx} 
\usepackage{natbib}  
\usepackage{caption} 
\usepackage{amsmath, amsthm, amssymb}
\usepackage[bottom]{footmisc}
\usepackage{booktabs}
\usepackage{color}
\usepackage{subfigure}
\usepackage[ruled, linesnumbered]{algorithm2e}
\usepackage{pdfpages}
\usepackage{threeparttable}
\usepackage{tablefootnote}
\usepackage[switch]{lineno}

\newcommand*\patchAmsMathEnvironmentForLineno[1]{%
  \expandafter\let\csname old#1\expandafter\endcsname\csname #1\endcsname
  \expandafter\let\csname oldend#1\expandafter\endcsname\csname end#1\endcsname
  \renewenvironment{#1}%
     {\linenomath\csname old#1\endcsname}%
     {\csname oldend#1\endcsname\endlinenomath}}%
\newcommand*\patchBothAmsMathEnvironmentsForLineno[1]{%
  \patchAmsMathEnvironmentForLineno{#1}%
  \patchAmsMathEnvironmentForLineno{#1*}}%
\AtBeginDocument{%
\patchBothAmsMathEnvironmentsForLineno{equation}%
\patchBothAmsMathEnvironmentsForLineno{align}%
\patchBothAmsMathEnvironmentsForLineno{flalign}%
\patchBothAmsMathEnvironmentsForLineno{alignat}%
\patchBothAmsMathEnvironmentsForLineno{gather}%
\patchBothAmsMathEnvironmentsForLineno{multline}%
}

\title{Neural Relational Inference with Efficient Message Passing Mechanisms}
\author{
    Siyuan Chen, Jiahai Wang\footnote{Corresponding author}, Guoqing Li
}
\affiliations{

    School of Computer Science and Engineering, Sun Yat-sen University, Guangzhou, China\\
    \{chensy47, ligq7\}@mail2.sysu.edu.cn, wangjiah@mail.sysu.edu.cn
}

\begin{document}
\maketitle


\begin{abstract}
    Many complex processes can be viewed as dynamical systems of interacting agents. In many cases, only the state sequences of individual agents are observed, while the interacting relations and the dynamical rules are unknown. The neural relational inference (NRI) model adopts graph neural networks that pass messages over a latent graph to jointly learn the relations and the dynamics based on the observed data. However, NRI infers the relations independently and suffers from error accumulation in multi-step prediction at dynamics learning procedure. Besides, relation reconstruction without prior knowledge becomes more difficult in more complex systems. This paper introduces efficient message passing mechanisms to the graph neural networks with structural prior knowledge to address these problems. A relation interaction mechanism is proposed to capture the coexistence of all relations, and a spatio-temporal message passing mechanism is proposed to use historical information to alleviate error accumulation. Additionally, the structural prior knowledge, symmetry as a special case, is introduced for better relation prediction in more complex systems. The experimental results on simulated physics systems show that the proposed method outperforms existing state-of-the-art methods.
\end{abstract}

\section{Introduction}
Many complex processes in natural and social areas including multi-agent systems \cite{Population,EvolveGraph}, swarm systems \cite{SwarmIntel}, physical systems \cite{DataDivenComplex,Glassy} and social systems \cite{AdaptiveSocial,SocialSystem} can be viewed as dynamical systems of interacting agents. Revealing the underlying interactions and dynamics can help us understand, predict, and control the behavior of these systems. However, in many cases, only the state sequences of individual agents are observed, while the interacting relations and the dynamical rules are unknown.

\begin{figure}
    \centering
    \includegraphics[width=0.95\columnwidth]{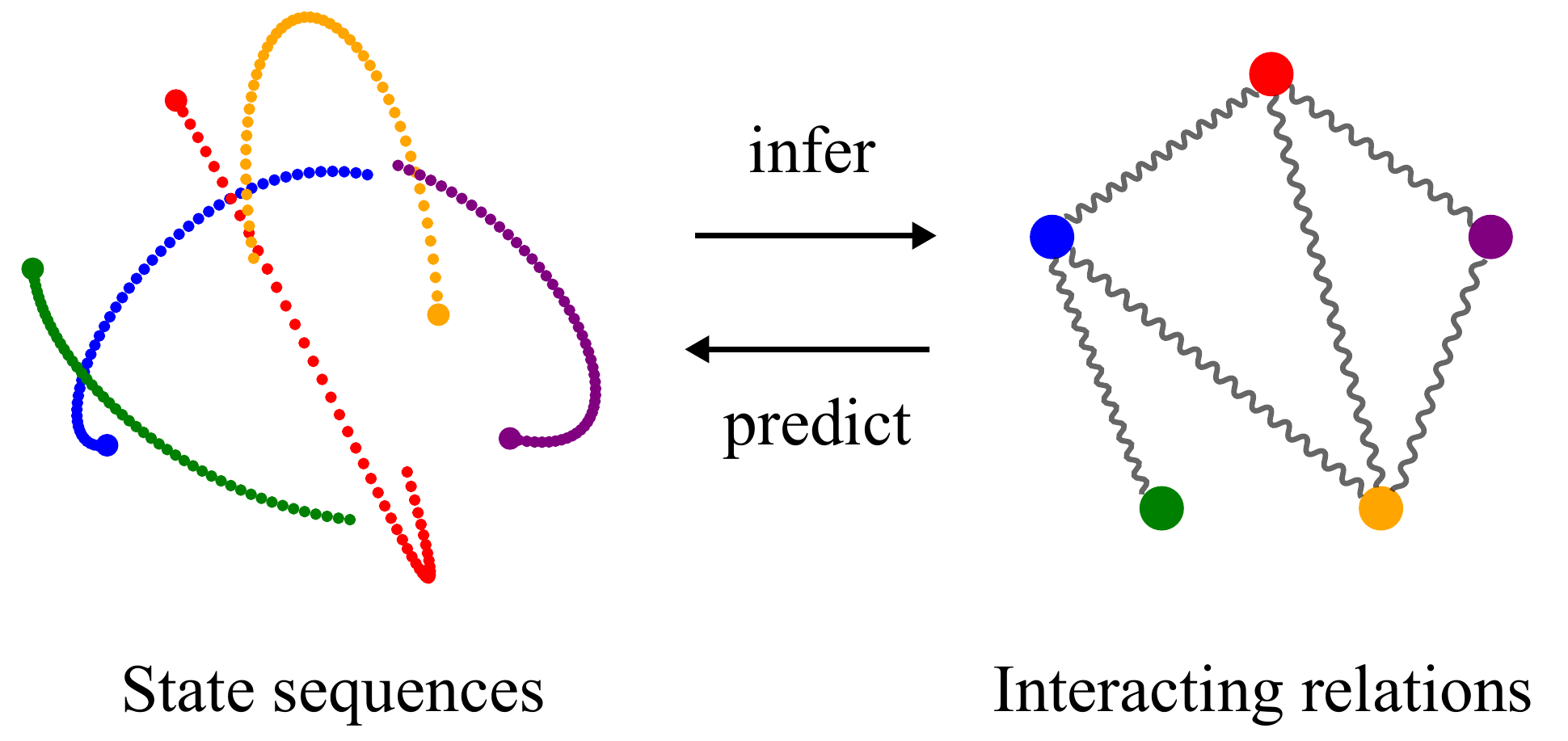}
    \caption{A dynamical system consisting of 5 particles that are linked by invisible springs. The interacting relations and future states are to be predicted based on the observed state sequences.}
    \label{fig:example}
\end{figure}

A lot of works \cite{Vain, RelationalEM, VisualNet} use implicit interaction models to learn the dynamics. These models can be regarded as graph neural networks (GNNs) over a fully-connected graph, and the implicit interactions are modeled via message passing operations \cite{VisualNet} or attention mechanisms \cite{Vain}. Compared with modeling implicit interactions, modeling the explicit interactions offers a more interpretable way to understand the dynamical systems. A motivating example is shown in Fig.~\ref{fig:example}~\cite{NRI}, where a dynamical system consists of 5 particles linked by invisible springs. It is of interest to infer the relations among the particles and predict their future states. Kipf et al.~(\citeyear{NRI}) propose the neural relational inference (NRI) model, a variational auto-encoder (VAE) \cite{VAE}, to jointly learn explicit interactions and dynamical rules in an unsupervised manner.

Currently, there are three main limitations of NRI. First, the interacting relations are inferred independently, while the coexistence of these relations is not considered. Alet et al.~(\citeyear{ModularMeta}) tackle this problem by taking all relations as a whole and iteratively improving the prediction through modular meta-learning. However, this method is computationally costly and it is limited to small-scale systems. Second, NRI predicts multiple steps into the future to emphasize the effect of the interactions, which leads to error accumulation and prevents the model from reconstructing the dynamics accurately. Third, as the scale and the complexity of systems increase, it is difficult to infer the interactions solely based on the observed data, while incorporating some structural prior knowledge can help better explore the structural space and promote the precision of relation recovery.

To address the problems above, this paper introduces efficient message passing mechanisms with structural prior knowledge for neural relational inference. For relation reconstruction, a relation interaction mechanism is introduced to capture the dependencies among different relations. For dynamics reconstruction, a spatio-temporal message passing mechanism is introduced to utilize historical information to alleviate the error accumulation of multi-step prediction. Additionally, the prior knowledge about the relations is incorporated as a regularization term in the loss function to impose soft constraints, and as a special case, the symmetry of relations is taken into consideration.

The contributions of this work are summarized as follows. 

\begin{itemize}
    \item Efficient message passing mechanisms are introduced to make a joint prediction of relations and alleviate the error accumulation of multi-step state prediction.
    \item The prior knowledge about relations, symmetry as a special case, is introduced to better reconstruct relations on more complex dynamical systems.
    \item Extensive experiments on physics simulation datasets are conducted. The results show the superiority of our method by comparing it with the state-of-the-art methods.
\end{itemize}

\section{Background}
\subsection{Neural Relational Inference (NRI)}
NRI \cite{NRI} is an unsupervised model that infers interacting structure from observational data and learns system dynamics. NRI takes the form of VAE, where the encoder infers the interacting relations, and the decoder predicts the future states of individual agents. 

Specifically, the encoder adopts GNNs with multiple rounds of message passing, and infers the distribution of potential interactions based on the input state,
\begin{align}
\mathbf{h} &= f_{\text{enc}}(\mathbf{x}), \\
q_\phi (\mathbf{z}|\mathbf{x}) &= \text{softmax}(\mathbf{h}),
\end{align}
where $\mathbf{x}=(\mathbf{x}_1^{1:T},...,\mathbf{x}_N^{1:T})$ is the observed trajectories of $N$ objects in the system in $T$ time steps, $f_{\text{enc}}$ is a GNN acting on the fully-connected graph, and $q_\phi (\mathbf{z}|\mathbf{x})$ is the factorized distribution of edge type $\mathbf{z}$.

Since directly sampling edges from $q_\phi (\mathbf{z}|\mathbf{x})$, a discrete distribution, is a non-differentiable process, the back-propagation cannot be used. To solve this problem, NRI uses the Gumbel-Softmax trick \cite{Gumbel-softmax}, which simulates a differentiable sampling process from a discrete distribution using a continuous function, i.e., 
\begin{equation}\label{eq:gumbel}
\mathbf{z}_{ij}=\text{softmax}\left((\mathbf{h}_{ij}+\mathbf{g})/\tau \right),
\end{equation}
where $\mathbf{z}_{ij}$ represents the edge type between the nodes $v_i$ and $v_j$, $\mathbf{g} \in \mathbb{R}^K$ is a vector of i.i.d. samples from a $\text{Gumbel}(0,1)$ distribution and the temperature $\tau$ is a parameter controlling the ``smoothness'' of the samples.

According to the inferred relations and past state sequences, the decoder uses another GNN that models the effect of interactions to predict the future state,
\begin{equation}
p_{\theta }(\mathbf{x}|\mathbf{z})= {\textstyle \prod_{t=1}^{T}}p_{\theta }(\mathbf{x}^{t+1}|\mathbf{x}^{1:t},\mathbf{z}),
\end{equation}
where $p_{\theta }(\mathbf{x}^{t+1}|\mathbf{x}^{1:t},\mathbf{z})$ is the conditional likelihood of $\mathbf{x}^{t+1}$.

As a variational auto-encoder model, NRI is trained to maximize the evidence lower bound,
\begin{equation}\label{eq:elbo}
\mathcal{L} = \mathbb{E}_{q_{\phi }(\mathbf{z}|\mathbf{x})}[{\rm{log}}\ p_{\theta }(\mathbf{x}|\mathbf{z})]-\text{KL}[q_{\phi }(\mathbf{z}|\mathbf{x})||p_{\theta }(\mathbf{z})],
\end{equation}
where the prior $p_{\theta }(\mathbf{z})= {\textstyle \prod_{i\ne j}}p_{\theta }(\mathbf{z}_{ij})$ is a factorised uniform distribution over edge types. In the right-hand side of Eq.~(\ref{eq:elbo}), the first term is the expected reconstruction error, while the second term encourages the approximate posterior $q_{\phi }(\mathbf{z}|\mathbf{x})$ to approach the prior $p_{\theta }(\mathbf{z})$.

\subsection{Message Passing Mechanisms in GNNs}
GNNs are a widely used class of neural networks that operates on graph structured data by message passing mechanisms \cite{MPNN}. For a graph $\mathcal{G=(V,E)}$ with vertices $v \in \mathcal{V}$ and edges $e=(v,v') \in \mathcal{E}$, the node-to-edge ($n\!\to\!e$) and edge-to-node ($e\!\to\!n$) message passing operations are defined as follows,
\begin{align}
v\!\rightarrow\!e&: & \mathbf{h}_{(i, j)}^{l} &= f_e^l\left(
\left[\mathbf{h}_i^l, \mathbf{h}_j^l, \mathbf{r}_{(i, j)}\right]
\right),\label{eq:v2e} \\ 
e\!\to\!v&: & \mathbf{h}_{j}^{l+1} &= {\textstyle f_v^l\left(
\left[\sum_{i\in\mathcal{N}_j}\mathbf{h}_{(i, j)}^l, \mathbf{x}_{j}\right]
\right)},\label{eq:e2v}
\end{align}
where $\mathbf{h}_i^l$ is the embedding of node $v_i$ in the $l$-th layer of the GNNs, $\mathbf{r}_{ij}$ is the feature of edge $e_{ij}$ (e.g. edge type), and $\mathbf{h}_{ij}^l$ is the embedding of edge $e_{ij}$ in the $l$-th layer of the GNNs. $\mathcal{N}_{j}$ denotes the set of indices of neighboring nodes with an incoming edge connected to vertex $v_j$, and $[\cdot ,\cdot]$ denotes the concatenation of vectors. $f_e$ and $f_v$ are node-specific and edge-specific neural networks, such as multilayer perceptrons (MLP),  respectively. Node embeddings are mapped to edge embeddings in Eq.~(\ref{eq:v2e}) and vice versa in Eq.~(\ref{eq:e2v}).

\section{Method}
\begin{figure}[!htbp]
    \centering
    \includegraphics[width=0.845\columnwidth]{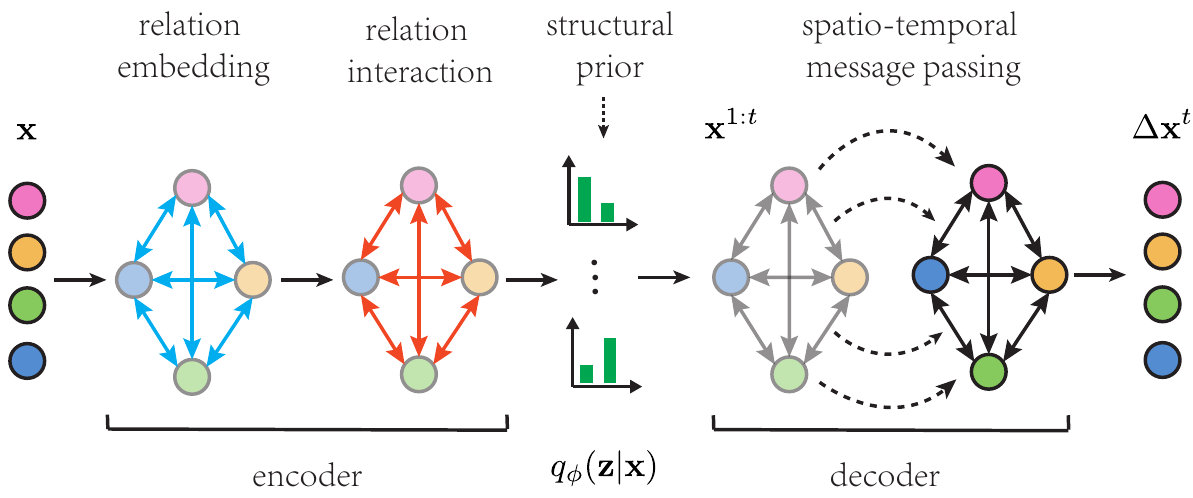}
    \caption{Overview of NRI-MPM. The encoder $q_\phi\left(\mathbf{z}|\mathbf{x}\right)$ uses the state sequences $\mathbf{x}$ to generate relation embeddings, and applies a relation interaction mechanism to jointly predict the relations. The structural prior knowledge of symmetry is imposed as a soft constraint for relation prediction. The decoder takes the predicted relations $\mathbf{z}$ and the historical state sequences $\mathbf{x}^{1:t}$ to predict the change in state $\Delta \mathbf{x}^t$.} 
    \label{fig:overview}
\end{figure}

The structure of our method is shown in Fig.~\ref{fig:overview}. Our method follows the framework of VAE in NRI. In the encoder, a relation interaction mechanism is used to capture the dependencies among the latent edges for a joint relation prediction. In the decoder, a spatio-temporal message passing mechanism is used to incorporate historical information for alleviating the error accumulation in multi-step prediction. As both mechanisms mentioned above can be regarded as Message Passing Mechanisms in GNNs for Neural Relational Inference, our method is named as NRI-MPM. Additionally, the structural prior knowledge, symmetry as a special case, is incorporated as a regularization term in the loss function to improve relation prediction in more complex systems. 

\subsection{Relation Interaction Mechanism}
The encoder is aimed at inferring the edge types $\mathbf{z}_{ij}$ based on the observed state sequences $\mathbf{x}=\left(\mathbf{x}^1,\dots,\mathbf{x}^T\right)$. From the perspective of message passing, the encoder defines a node-to-edge message passing operation at a high level. As shown in Eqs.~(\ref{eq:nri_enc_1})-(\ref{eq:nri_enc_4}), NRI first maps the observed state $\mathbf{x}_j$ to a latent vector $\mathbf{h}^1_j$, and then applies two rounds of node-to-edge and one round of edge-to-node message passing alternately to obtain the edge embeddings $\mathbf{h}_{(i, j)}^2$ that integrate both local and global information. Then, $\mathbf{h}_{(i, j)}^2$ are used to predict the pairwise interacting types independently.  
\begin{align}
    & & \mathbf{h}_j^1 &= f_{\text{emb}}(\mathbf{x}_j),\label{eq:nri_enc_1} \\
    v\!\to\!e:& & \mathbf{h}_{(i, j)}^1 &= f^1_e([\mathbf{h}^1_i, \mathbf{h}^1_j]), \\
    e\!\to\!v:& & \mathbf{h}_j^2 &= {\textstyle f^1_v\left(\sum_{i\neq j}\mathbf{h}_{(i, j)}^1\right)},\\
    v\!\to\!e:& & \mathbf{h}_{(i, j)}^2 &= f^2_e([\mathbf{h}^2_i, \mathbf{h}^2_j]).\label{eq:nri_enc_4}
\end{align}

However, the relations are generally dependent on each other since they jointly affect the future states of individual agents. Within the original formulations of GNNs, Eqs.~(\ref{eq:nri_enc_1})-(\ref{eq:nri_enc_4}) cannot effectively modeled the dependencies among edges. Typical GNNs are designed to learn node embeddings, while the edge embeddings are treated as a transient part of the computation. To capture the coexistence of all relations, this paper introduces an edge-to-edge ($e\!\to\!e$) message passing operation that directly passes messages among edges, named the relation interaction mechanism,
\begin{equation}
    \resizebox{0.98\columnwidth}{!}{$
    e\!\to\!e:\left\{\mathbf{e}_{(i, j)}:(v_i, v_j)\in \mathcal{E}\right\}
                = g_e\left(\left\{\mathbf{h}^2_{(i, j)}:(v_i, v_j)\in \mathcal{E}\right\}\right).
    $}
\end{equation}

\begin{figure}[!htbp]
    \centering
    \includegraphics[width=0.92\columnwidth]{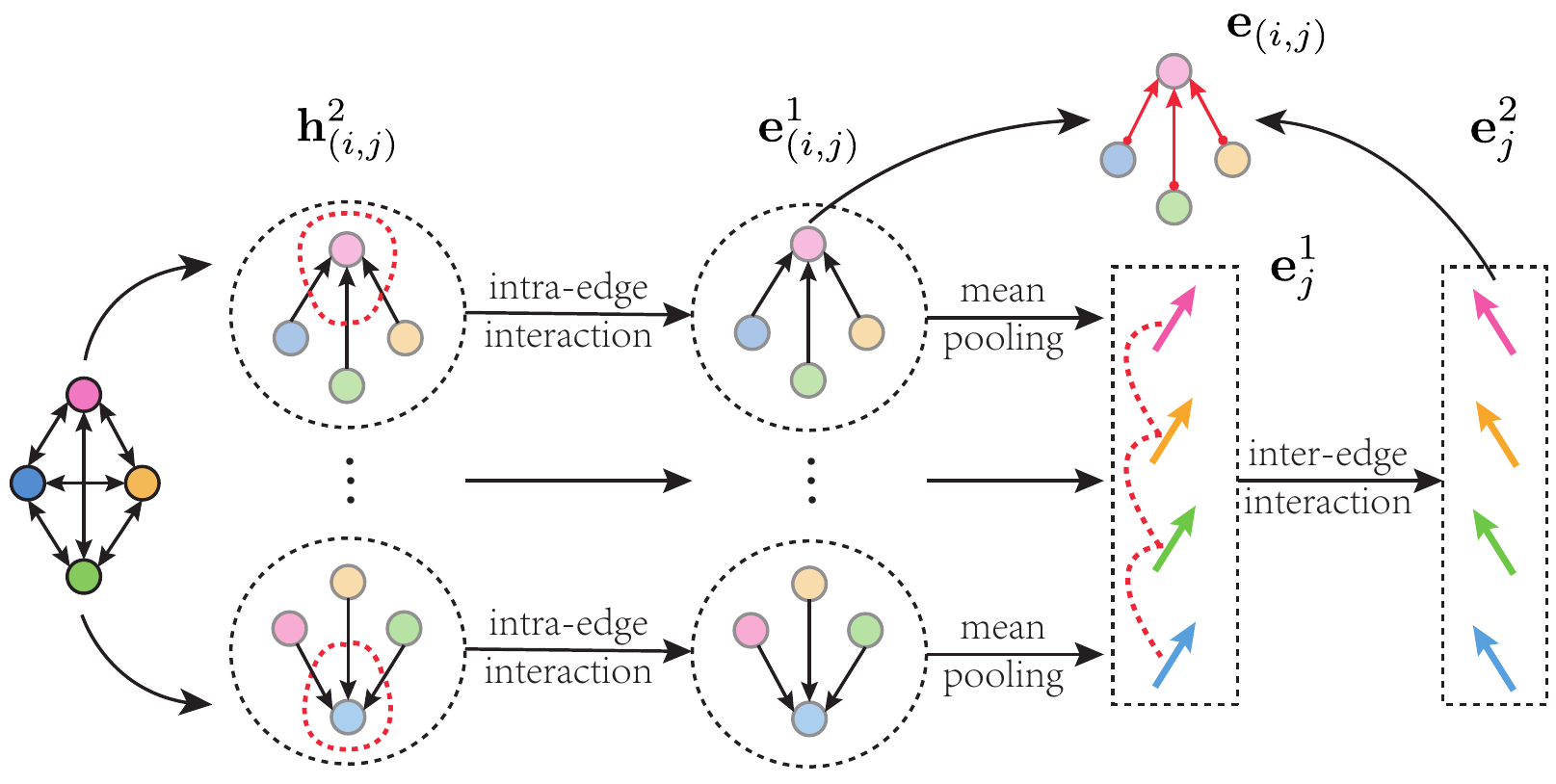}
    \caption{Relation interaction mechanism. Given the edge embeddings $\mathbf{h}_{(i, j)}^2$, the intra-edge and inter-edge interaction operations are used to model the interactions among incoming edges of the same node and those among the incoming edges of different nodes, respectively. The resulting embeddings $\mathbf{e}^1_{(i, j)}$ and $\mathbf{e}^2_{j}$ are concatenated to obtain the final edge representations $\mathbf{e}_{(i, j)}$.}
    \label{fig:relation interaction}
\end{figure}

Ideally, this operation includes modeling the pairwise dependencies among all edges, which is computationally costly as its time complexity is $O(|\mathcal{E}|^2)$. Alternatively, as shown in Fig.~\ref{fig:relation interaction}, our method decomposes this operation into two sub-operations, intra-edge interaction and inter-edge interaction operations, for modeling the interactions among incoming edges of the same node and those among the incoming edges of different nodes, respectively. 

The intra-edge interaction operation is defined as follows, 
\begin{equation}
    e\!\to\!e:\left\{\mathbf{e}^1_{(i, j)}:i\in \mathcal{N}_j\right\}
                = g^{\text{intra}}_e\left(\left\{\mathbf{h}^2_{(i, j)}:i\in \mathcal{N}_j\right\}\right).
\end{equation}
{From the above definition, $g^{\text{intra}}_e$ is required to be permutation equivariant \cite{SetTransformer} to preserve the correspondences between the input and output edge embeddings. Formally, let $\mathcal{I}_j=\left\{i_k\right\}^{|\mathcal{N}_j|}_{k=1}$ be the ascending sequence of $\mathcal{N}_j$ and $\mathcal{S}_j$ be the set of all permutations on $\mathcal{I}_j$. For any permutation $\pi\in\mathcal{S}_j$, $\pi(\mathbf{h}^2_{(\mathcal{I}_j, j)})=\mathbf{h}^2_{(\pi(\mathcal{I}_j), j)}$. $g^{\text{intra}}_e$ is said to be permutation equivariant if $\mathbf\pi({e}^1_{(\mathcal{I}_j, j)}) = g^{\text{intra}}_e\circ\pi(\mathbf{h}^2_{(\mathcal{I}_j, j)})$, where $\circ$ denotes the composition of functions. Although permutation invariant operators such as sum and max aggregators are widely used in GNNs \cite{HowPowerful}, the design for permutation equivariant operators is less explored. Inspired by Hamilton et al.~(\citeyear{GraphSage}), this paper treats a set as an unordered sequence and defines $g^{\text{intra}}_e$ as a sequence model over it, appending by an inverse permutation to restore the order, i.e.,}
\begin{align}
    e\!\to\!e:\quad
    \mathbf{e}^1_{(\mathcal{I}_j, j)} = \pi^{-1}\circ S\circ\pi\left(\mathbf{h}^2_{(\mathcal{I}_j, j)}\right),
    \label{eq:intra_0}
\end{align}
where $S$ is a sequence model such as recurrent neural networks (RNNs) and convolutional neural networks (CNNs). {When $S$ is implemented by self-attention, $g^{\text{intra}}_e$ is permutation equivariant, which may not hold when $S$ is an RNN. In this case, $g^{\text{intra}}_e$ is expected to be approximately permutation equivariant with a well trained $S$.} 

Similarly, one can define an inter-edge interaction operation. The difference is that this operation treats all incoming edges of a node as a whole. For simplicity, this paper applies mean-pooling to the edge embeddings to get an overall representation. Formally, the inter-edge interaction operation is defined to be the composition of the following steps,
\begin{align}
    e\!\to\!v:& & \mathbf{e}^1_{j} &= \text{Mean}\left(\mathbf{e}^1_{(1:N, j)}\right),\label{eq:inter_0} \\
    v\!\to\!v:& & \left\{\mathbf{e}^2_j\right\}^N_{j=1} &= g^{\text{inter}}_e\left(\left\{\mathbf{e}^1_j\right\}^N_{j=1}\right),\label{eq:inter_1}
\end{align}
where $g^{\text{inter}}_e$ is a node-to-node ($v\!\to\!v$) operation that passes messages among nodes and takes a similar form of $g^{\text{intra}}_e$. {To analyze the time complexity of relation interaction, this paper assumes that all sequence models involved are RNNs.} Since $g^{\text{intra}}_e$ and $\text{Mean}$ can be applied to each node in parallel and the time complexity of Eqs.~(\ref{eq:intra_0})-(\ref{eq:inter_1}) are all $O(N)$, the overall time complexity of relation interaction is $O(N)$, which is much more effective than the pairwise interaction.

Finally, an MLP is used to unify the results of the two operations, and the predicted edge distribution is defined as
    \begin{align}\label{eq:enc}
        \mathbf{e}_{(i, j)} &= \text{MLP}([\mathbf{e}^1_{(i, j)}, \mathbf{e}^2_{j}]),\\
        q_\phi (\mathbf{z}_{ij}|\mathbf{x}) &= \text{softmax}(\mathbf{e}_{(i,j)}).
    \end{align}
As in Eq.~(\ref{eq:gumbel}), $\mathbf{z}_{ij}$ is sampled via the Gumbel-Softmax trick to allow back-propagation.

\subsection{Spatio-Temporal Message Passing Mechanism }

The decoder is aimed at predicting the future state $\mathbf{x}^{t+1}$ using the inferred relations $\mathbf{z}$ and the historical states $\mathbf{x}^{1:t}$. 
Since the interactions can have a small effect on short-term dynamics, NRI predicts multiple steps into the future to avoid a degenerate decoder that ignores the effect of interactions. In multi-step prediction, the predicted value $\boldsymbol\mu_{j}^{t}$ is replaced by the ground truth state $\mathbf{x}_j^{t}$, i.e.,
\begin{align}
    \boldsymbol\mu_{j}^{t+1} &= f_{\text{dec}}(\mathbf{x}_j^t), \\
    \boldsymbol\mu_{j}^{t+m} &= f_{\text{dec}}(\boldsymbol\mu_{j}^{t+m-1}),\qquad \qquad m=2,\dots,M
\end{align}
where $f_{\text{dec}}$ denotes the decoder, and $M (M\geq 1)$ is the number of time steps to predict. This means that errors in the prediction process accumulate over $M$ steps. As the decoder of NRI only uses the current state to predict future states, it is difficult to avoid error accumulation. Apparently, the current state of the interacting system is related to the previous states, and thus, incorporating historical information can help learn the dynamical rules and alleviate the error accumulation in multi-step prediction. To this end, sequence models are used to capture the non-linear correlation between previous states and the current state. GNNs combined with sequence models can be used to capture spatio-temporal information, which has been widely used in traffic flow forecasting \cite{DCRNN, ASTGCN}. In this paper, the sequence model can be composed of one or more of CNNs, RNNs, attention mechanisms, etc.

Inspired by interaction networks \cite{Interaction-Networks} and graph networks \cite{GNs-block}, sequence models are added to the original node-to-edge and edge-to-node operations to obtain their spatio-temporal versions. In this way, the decoder can integrate the spatio-temporal interaction information at different fine-grained levels, i.e., the node level and the edge level, which can help better learn the dynamical rules. 

\begin{figure}[!htbp]
    \centering
    \includegraphics[width=0.99\columnwidth]{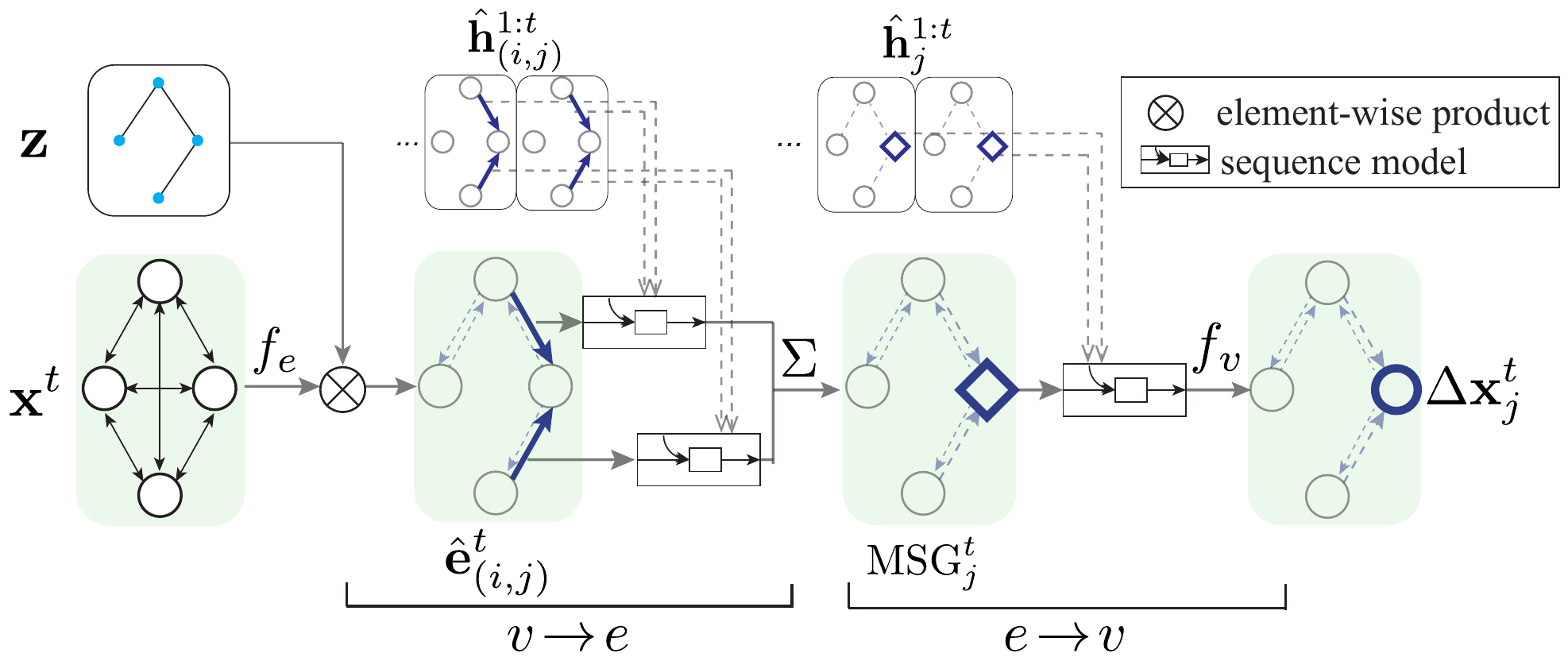}
    \caption{The structure of the decoder. The decoder takes the interacting relations $\mathbf{z}$, the current state $\mathbf{x}^t$, the historical hidden states $\hat{\mathbf{h}}^{1:t}_{(i, j)}$ and $\hat{\mathbf{h}}^{1:t}_j$ as inputs to predict the change in state $\Delta \mathbf{x}^t$. Sequence models are added to the message passing operations to jointly capture the spatio-temporal dependencies. Elements that are currently updated are highlighted in blue, e.g., $\hat{\mathbf{e}}_{(i,j)}^t$ is an edge embedding updated in the node-to-edge ($v\!\to\!e$) message passing operation.}
    \label{fig:dec}
\end{figure}

As shown in Fig.~\ref{fig:dec}, the decoder contains a node-to-edge and an edge-to-node spatio-temporal message passing operations. The node-to-edge spatio-temporal message passing operation is defined as
\begin{align}\label{eq:seqedge}
    v\!\to\!e:\quad\hat{\mathbf{e}}_{(i,j)}^t &= \sum_{k}z_{ij,k}\hat{f}_e^k ([\mathbf{x}_i^t,\mathbf{x}_j^t]), \\
    \hat{\mathbf{h}}_{(i,j)}^{t+1} &= S_{\text{edge}}\left([\hat{\mathbf{e}}_{(i,j)}^t,\hat{\mathbf{h}}_{(i,j)}^{1:t}]\right),
\end{align}
where $z_{ij,k}$ is the $k$-th element of the vector $\mathbf{z}_{ij}$, representing the edge type $k$, whose effect is modeled by an MLP $\hat{f}_e^k$. For each edge $e_{ij}$, the effect of all potential interactions are aggregated into $\hat{\mathbf{e}}_{(i,j)}^t$ as a weighted sum. Its concatenation with the previous hidden states $\hat{\mathbf{h}}_{(i,j)}^{1:t}$ is fed to $S_{\text{edge}}$ to generate the future hidden state of interactions $\hat{\mathbf{h}}_{(i,j)}^{t+1}$ that captures the temporal dependencies at edge level.

Similarly, the edge-to-node spatio-temporal message passing operation is defined as
\begin{align}\label{eq:seqnode}
e\!\to\!v:\text{MSG}_j^t &= {\textstyle \sum_{i\ne j}\hat{\mathbf{h}}_{(i,j)}^{t+1}}, \\
\hat{\mathbf{h}}_{j}^{t+1} &= S_{\text{node}}\left([\text{MSG}_j^t,\mathbf{x}_j^t],\hat{\mathbf{h}}_{j}^{1:t}\right).
\end{align}
For each node $v_j$, the spatial dependencies $\hat{\mathbf{h}}_{(i,j)}^{t+1}$ are aggregated into $\text{MSG}_j^t$. $\text{MSG}_j^t$ together with the current state $\mathbf{x}_j^t$ and the previous hidden states $\hat{\mathbf{h}}_{j}^{1:t}$ are fed to $S_{\text{node}}$ to generate the future hidden state of nodes $\hat{\mathbf{h}}_{j}^{t+1}$ that captures the temporal dependencies at node level.

Finally, the predicted future state is defined as
\begin{align}\label{eq:dec}
    \Delta \mathbf{x}_{j}^{t} &= \hat{f}_v\left(\hat{\mathbf{h}}_{j}^{t+1}\right), \\
    \boldsymbol\mu_{j}^{t+1} &= \mathbf{x}_j^t + \Delta \mathbf{x}_j^{t}, \\
    p_{\theta}(\mathbf{x}^{t+1}|\mathbf{x}^{1:t},\mathbf{z}) &= \mathcal{N}(\mathbf{x}^{t+1}|\boldsymbol\mu^{t+1},\sigma ^2\mathbf{I}),
\end{align}
where $\hat{f}_v$ is an MLP predicting the change in state $\Delta \mathbf{x}_j^{t}$, and $\sigma^2$ is a fixed variance. Note that since both $\hat{\mathbf{h}}^{1:t}_{(i, j)}$ and $\hat{\mathbf{h}}^{1:t}_j$ rely on the historical states $\mathbf{x}^{1:t}$, the decoder is implicitly a function of $\mathbf{x}^{1:t}$.

\subsection{Structural Prior Knowledge}
As the scale and complexity of systems increase, it becomes more difficult to infer relations solely based on the state sequences of individual agents. Therefore, it is desirable to incorporate possible structural prior knowledge, such as sparsity \cite{NRI}, symmetry and node degree distribution \cite{SUGAR}. Since symmetric relations widely exist in physical dynamical systems, the symmetry of relations is studied as a special case. Li et al.~(\citeyear{SUGAR}) impose a hard constraint on symmetry, i.e., setting $\mathbf{z}_{ji}=\mathbf{z}_{ij}$. However, the hard constraint may limit the exploration of the model during the training procedure, and sometimes it may lead to a decrease of the prediction precision. By contrast, this paper imposes a soft constraint by adding a regularization term to the original loss function.

Specifically, an auxiliary distribution $q'_{\phi}(\mathbf{z}|\mathbf{x})$ is introduced as the ``transpose'' of the predicted relation distribution $q_{\phi}(\mathbf{z}|\mathbf{x})$, namely,
\begin{equation}\label{eq:transpose}
    q'_{\phi}(\mathbf{z}_{ij}|\mathbf{x}) = q_{\phi}(\mathbf{z}_{ji}|\mathbf{x}).
\end{equation}
Then, the Kullback-Leibler divergence between $q'_{\phi}(\mathbf{z}|\mathbf{x})$ and $q_{\phi}(\mathbf{z}|\mathbf{x})$ is used as a regularization term for symmetry, i.e.,
\begin{equation}\label{eq:prior}
    \mathcal{L}' = \mathcal{L} - \lambda \cdot \text{KL}\left[q'_{\phi}(\mathbf{z}|\mathbf{x})||q_{\phi}(\mathbf{z}|\mathbf{x})\right],
\end{equation}
where $\lambda$ is a penalty factor for the symmetric prior. 

Notations used in this paper, details of the computation of $\mathcal{L}'$ and the pseudo code of NRI-MPM are shown in \textit{Appendix A, Appendix B and Appendix C}, respectively.

\section{Experiments}
All methods are tested on three types of simulated physical systems: particles connected by springs, charged particles and phase-coupled oscillators, named Springs, Charged and Kuramoto, respectively. {For the Springs and Kuramoto datasets, objects do or do not interact with equal probability. For the Charged datasets, objects attract or repel with equal probability.} For each type of system, a 5-object dataset and a 10-object dataset are simulated. All datasets contain 50k training samples, and 10k validation and test samples. Further details on data generation can be found in \cite{NRI}. All methods are evaluated w.r.t. two metrics, the accuracy of relation reconstruction and the mean squared error (MSE) of future state prediction.

For our method, the sequence models used in both encoders and decoders are composed of gated recurrent units (GRUs) \cite{GRU}, except that attention mechanisms are added to the decoder in the Kuramoto datasets (see \textit{Appendix D}). The penalty factor $\lambda$ is set to $10^2$ for all datasets except that it is set to 1 and $10^3$ for the 5-object and 10-object Kuramoto datasets, respectively (see \textit{Appendix E}).
\subsection{Baselines}
To evaluate the performance of our method, we compare it with several competitive methods as follows.
\begin{itemize}
  \item Correlation \cite{NRI}: a baseline that predicts the relations between two particles based on the correlation of their state sequences.
  \item LSTM \cite{NRI}: an LSTM that takes the concatenation of all state vectors and predict all future states simultaneously.
  \item NRI \cite{NRI}: the neural relational inference model that jointly learns the relations and the dynamics with a VAE.
  \item SUGAR \cite{SUGAR}: a method that introduces structural prior knowledge such as hard symmetric constraint and node degree distribution for relational inference.
  \item ModularMeta \cite{ModularMeta}: a method that solves the relational inference problem via modular meta-learning.
\end{itemize}

 To compare the performance of our method with ``gold standard'' methods, i.e., those trained given the ground truth relations, this paper introduces two variants as follows.

\begin{itemize}
  \item Supervised: a variant of our method that only trains an encoder with the ground truth relations.
  \item NRI-MPM (true graph): a variant of our method that only trains a decoder given the ground truth relations.
\end{itemize}

The codes of NRI\footnote{https://github.com/ethanfetaya/nri} and  ModularMeta\footnote{https://github.com/FerranAlet/modular-metalearning} are public and thus directly used in our experiments. {SUAGR is coded by ourselves according to the original paper.} Note that Correlation and Supervised are only designed for relation reconstruction, while LSTM and NRI-MPM (true graph) are only designed for state prediction.

 To verify the effectiveness of the proposed message passing mechanisms and the structural prior knowledge, this paper introduces some variants of our method as follows.
\begin{itemize}
  \item NRI-MPM w/o RI: a variant of our method without the relation interaction mechanism.
  \item NRI-MPM w/o intra-RI, NRI-MPM w/o inter-RI: variants of our method without the intra- and inter-edge interaction operations, respectively.
  \item NRI-MPM w/o ST: a variant of our method without the spatio-temporal message passing mechanism.
  \item NRI-MPM w/o Sym, NRI-MPM w/ hard Sym: variants of our method without the symmetric prior and that with hard symmetric constraint, respectively.
\end{itemize}
\subsection{Comparisons with Baselines}
\begin{table}[!t]
  \centering
  \caption{Accuracy (\%) of relation reconstruction.}
  \resizebox{0.95\columnwidth}{!}{
    \begin{threeparttable}
    \begin{tabular}{lccc}
    \toprule
    Model & Springs & Charged & Kuramoto \\
    \midrule
    \multicolumn{4}{c}{5 objects} \\
    Correlation   & $52.4\pm 0.0$ & $55.8\pm 0.0$ & $62.8\pm 0.0$ \\
    NRI   & $\mathbf{99.9}\pm 0.0$ & $82.1\pm 0.6$ & $96.0\pm 0.1$ \\
    SUGAR   & $\mathbf{99.9}\pm 0.0^*$ & $82.9\pm 0.8^*$ & $91.8\pm 0.1^*$ \\
    ModularMeta & $\mathbf{99.9}$     & $88.4$     & $96.2\pm 0.3^*$ \\
    NRI-MPM  & $\mathbf{99.9}\pm 0.0$     & $\mathbf{93.3}\pm 0.5$     & $\mathbf{97.3}\pm 0.2$ \\
    \midrule
    \midrule
    Supervised & $99.9\pm 0.0$     & $95.4\pm 0.1$     & $99.3\pm 0.0$ \\
    \midrule
    \multicolumn{4}{c}{10 objects} \\
    Correlation   & $50.4\pm 0.0$ & $51.4\pm 0.0$ & $59.3\pm 0.0$ \\
    NRI   & $98.4\pm 0.0$ & $70.8\pm 0.4$ & $75.7\pm 0.3$ \\
    SUGAR   & $98.3\pm 0.0^*$ & $72.0\pm 0.9^*$ & $74.0\pm 0.2^*$ \\
    ModularMeta & $98.8\pm 0.0^*$     & $63.8\pm 0.1^*$     & $\mathbf{89.6}\pm 0.1^*$ \\
    NRI-MPM  & $\mathbf{99.1}\pm 0.0$     & $\mathbf{81.6}\pm 0.2$     & $80.3\pm 0.6$ \\
    \midrule
    \midrule
    Supervised & $99.4\pm 0.0$     & $89.7\pm 0.1$     & $94.3\pm 0.8$ \\
    \bottomrule
    \end{tabular}%
  
  \begin{tablenotes}
    \footnotesize
    \item[*] The results in these datasets are unavailable in the original paper, and they are obtained by running the codes provided by the authors.
  \end{tablenotes}
\end{threeparttable}
  }
  \label{tab:acc}%
\end{table}%
\begin{table*}[!htbp]
  \centering
  \caption{Mean squared error in predicting future states for simulations with 5 interacting objects.}
  \resizebox{0.99\textwidth}{!}{
    \begin{threeparttable}
    \begin{tabular}{l|ccc|ccc|ccc}
    \toprule
        Datasets  & \multicolumn{3}{c|}{Springs} & \multicolumn{3}{c|}{Charged} & \multicolumn{3}{c}{Kuramoto} \\
    \midrule
    Predictions steps & \multicolumn{1}{c}{1} & \multicolumn{1}{c}{10} & \multicolumn{1}{c|}{20} & \multicolumn{1}{c}{1} & \multicolumn{1}{c}{10} & \multicolumn{1}{c|}{20} & \multicolumn{1}{c}{1} & \multicolumn{1}{c}{10} & \multicolumn{1}{c}{20} \\
    \midrule
    LSTM    & 4.13e-8     & 2.19e-5     & 7.02e-4     & 1.68e-3     & 6.45e-3     & 1.49e-2     & \textbf{3.44e-4}     & \textbf{1.29e-2}     & 4.74e-2 \\
    NRI   & 3.12e-8     & 3.29e-6     & 2.13e-5     & 1.05e-3     & 3.21e-3     & 7.06e-3     & 1.40e-2     & 2.01e-2     & \textbf{3.26e-2} \\
    SUGAR   & 3.71e-8$^*$     & 3.86e-6$^*$     & 1.53e-5$^*$     & 1.18e-3$^*$     & 3.43e-3$^*$     & 7.38e-3$^*$     & 2.12e-2$^*$     & 9.45e-2$^*$     & 1.83e-1$^*$ \\
    ModularMeta & 3.13e-8     & 3.25e-6     & -     & 1.03e-3     & 3.11e-3     & -     & 2.35e-2$^*$     & 1.10e-1$^*$     & 1.96e-1$^*$ \\
    NRI-MPM  & \textbf{8.89e-9}     & \textbf{5.99e-7}     & \textbf{2.52e-6}     & \textbf{7.29e-4}     & \textbf{2.57e-3}     & \textbf{5.41e-3}     & 1.57e-2     & 2.73e-2     & 5.36e-2 \\
    \midrule
    NRI-MPM (true graph) & 1.60e-9     & 9.06e-9     & 1.50e-7     & 8.06e-4     & 2.51e-3     & 5.66e-3     & 1.73e-2     & 2.49e-2     & 4.09e-2 \\
    \bottomrule
    \end{tabular}%
    \begin{tablenotes}
      \footnotesize
      \item[*] Results in these datasets are unavailable in the original paper, and they are obtained by running the codes provided by the authors.
    \end{tablenotes}
  \end{threeparttable}
  }
  \label{tab:mse}%
\end{table*}%
\begin{figure*}[t]
  \centering
\subfigure[LSTM]{
  \begin{minipage}{0.32\textwidth}
  \centering
  \includegraphics[width=\textwidth]{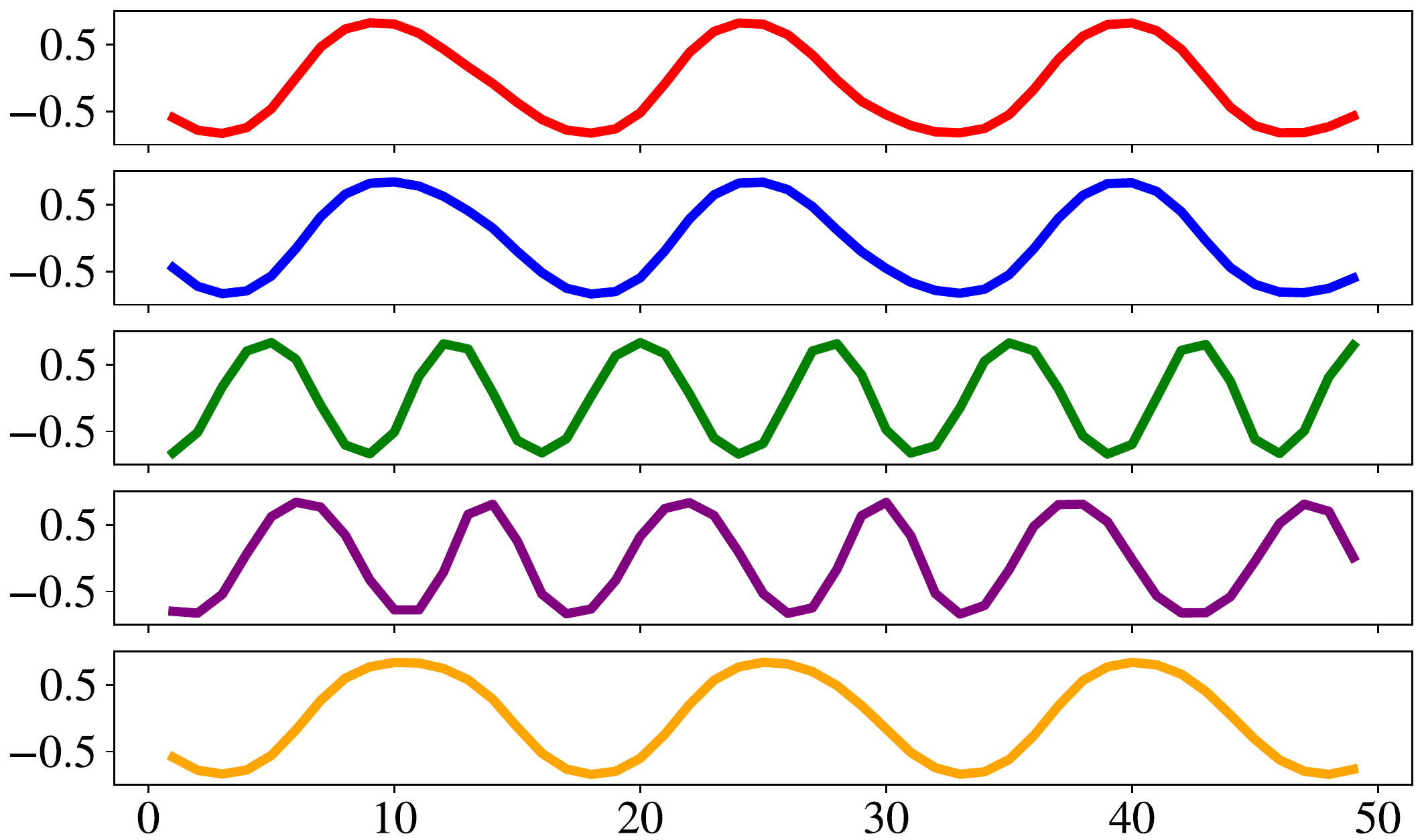}
  \end{minipage}
}
\subfigure[Ground Truth]{
  \begin{minipage}{0.32\textwidth}
  \centering
  \includegraphics[width=\textwidth]{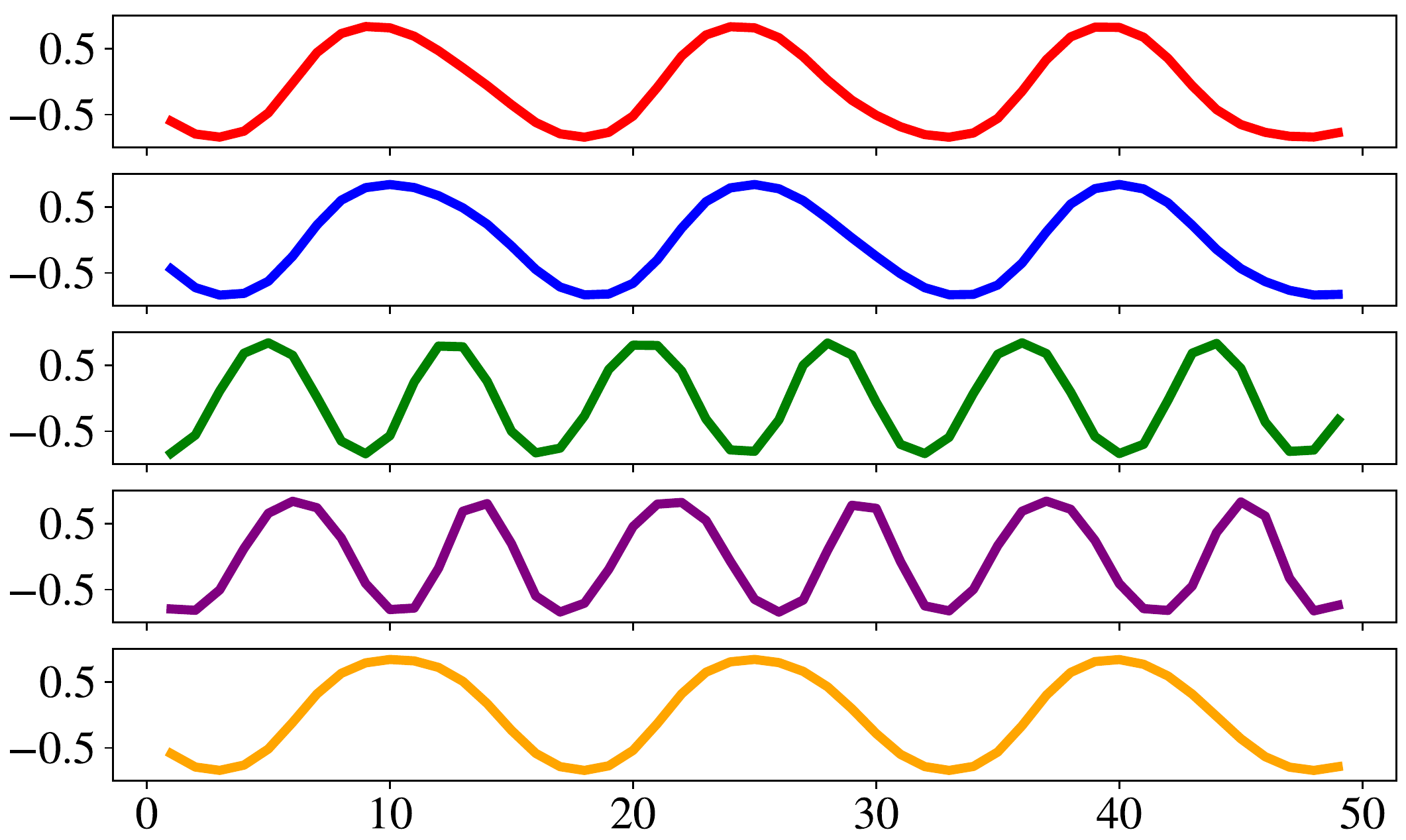}
  \end{minipage}
}
\subfigure[NRI-MPM]{
  \begin{minipage}{0.32\textwidth}
  \centering
  \includegraphics[width=\textwidth]{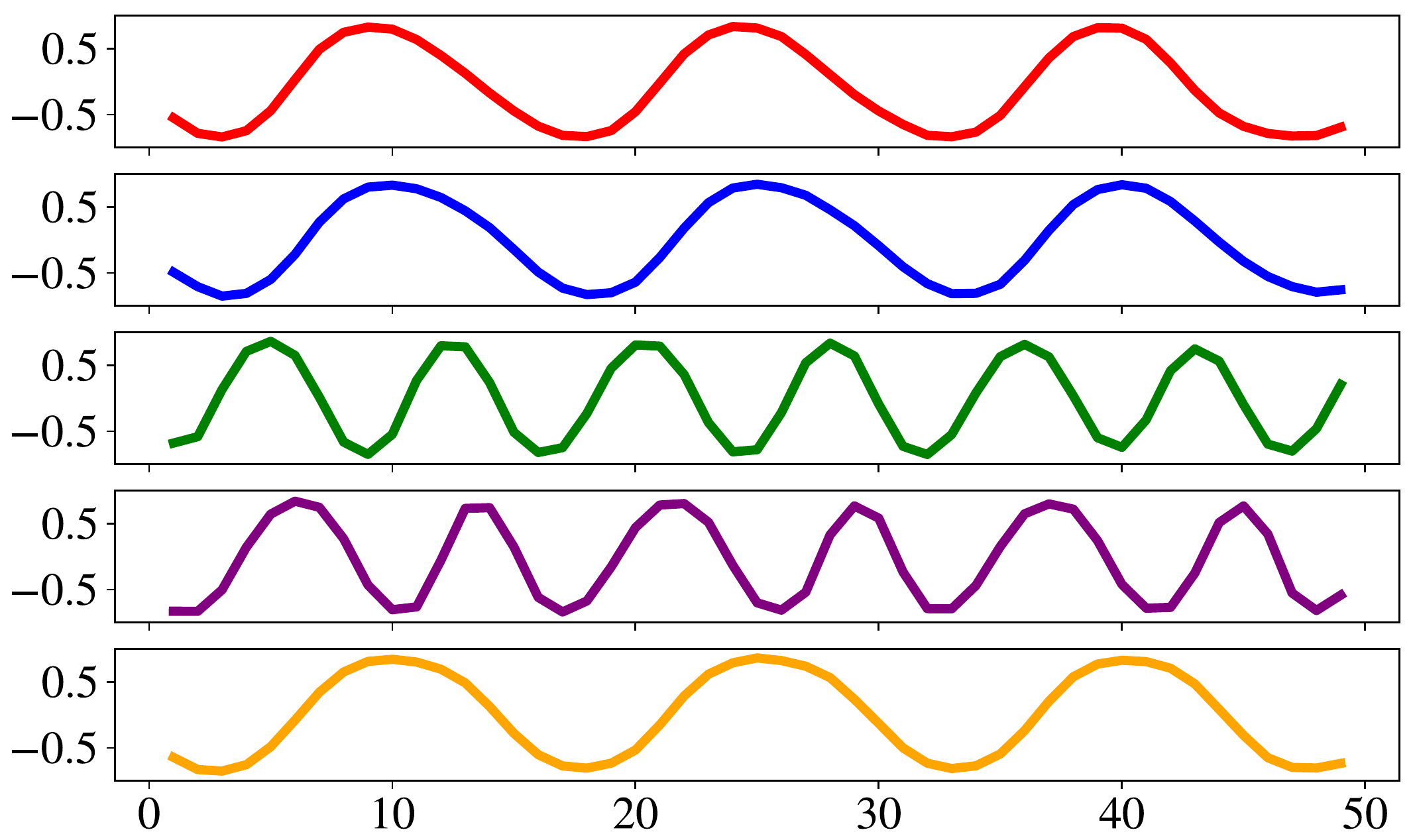}
  \end{minipage}
}
\caption{Visualization of the ground truth states (b) together with the predicted states for LSTM (a) and NRI-MPM (c) in the 5-object Kuramoto dataset.}
\label{fig:qualatitive}
\end{figure*}
The results of relation reconstruction and future state prediction in the 5-object datasets (for MSEs in the 10-object datasets, see \textit{Appendix E}) are shown is Table~\ref{tab:acc} and Table~\ref{tab:mse}, respectively. Our method significantly outperforms all baselines in nearly all datasets in terms of both accuracy and MSE. In the Springs datasets, the accuracies of NRI, SUGAR, ModularMeta and our method are all comparable with the supervised baseline, while our method achieves significantly lower MSEs in the 5-object systems. This indicates that our method can better learn the relations and dynamics simultaneously. In the Charged datasets, our method outperforms the baselines by 4.9\%-9.6\% in terms of accuracy. The reason may be that the charged systems are densely connected since a charged particle interacts with all other particles, and our method can better handle this situation. {SUGAR achieves higher accuracies than NRI, suggesting the effectiveness of structural prior knowledge. Still, our method performs better with the extra help of the proposed message passing mechanisms.} 

{In the Kuramoto datasets, the performance of our method is non-dominant. Our method achieves higher accuracies than NRI, while the MSEs are larger. ModularMeta achieves higher accuracy than our method in the 10-object system, while the MSEs are much poorer than all other methods. Maybe the interactions among objects in the Kuramoto dataset are relatively weak \cite{Kuramoto}, making it more difficult to infer the relations based on the observed states, and the situation worsens as the scale of systems increases. ModularMeta infers all relations as a whole, and modifies its predictions with the help of simulated annealing, which may help better search the structural space to achieve better relation predictions. However, this does not translate to better state prediction. According to \cite{Kuramoto}, Kuramoto is a synchronization system where the object states converge to a certain set of values in the long run. Maybe the interaction is less helpful for state prediction in this case.}
\begin{figure}[!t]
  \centering
  \includegraphics[width=0.85\columnwidth]{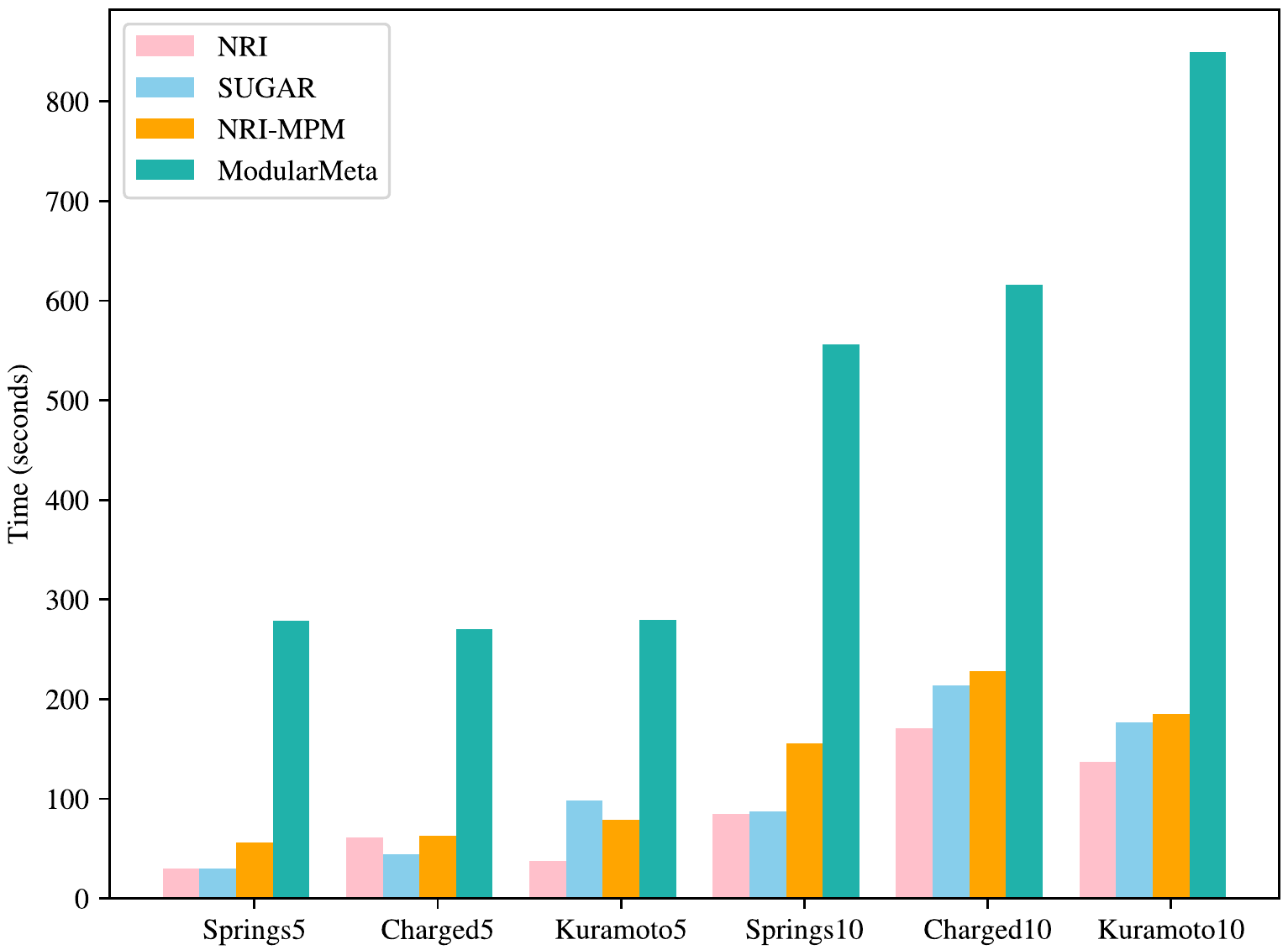}
  \caption{Running times of different methods.}
  \label{fig:runtime}
\end{figure}
\paragraph{Qualitative Analysis of Future State Prediction}
{One observes that LSTM achieves lower MSEs for short-term prediction in the 5-object Kuramoto dataset, but its performance declines for long-term prediction. To further understand the predictive behavior of LSTM and our method, we conduct a qualitative analysis by visualizing the predicted states together with the ground truth states in 49 steps, and the results are shown in Fig.~\ref{fig:qualatitive}. From Fig.~\ref{fig:qualatitive}(a), LSTM can capture the shape of the sinusoidal waveform but fails to make accurate prediction for time steps larger than 40 (e.g., curves in green and purple). By contrast, as shown in Fig.~\ref{fig:qualatitive}(c), the predicted states of our method closely match the ground truth states except for those in the last few time steps for the third particle, whose curve is colored in green. Maybe our method can better capture the interactions among particles that affect the long-term dynamics. Note that this result is consistent with that reported in \textit{Appendix A.1} in the original paper of NRI.}

\begin{figure}[!t]
  \centering
  \includegraphics[width=0.8\columnwidth]{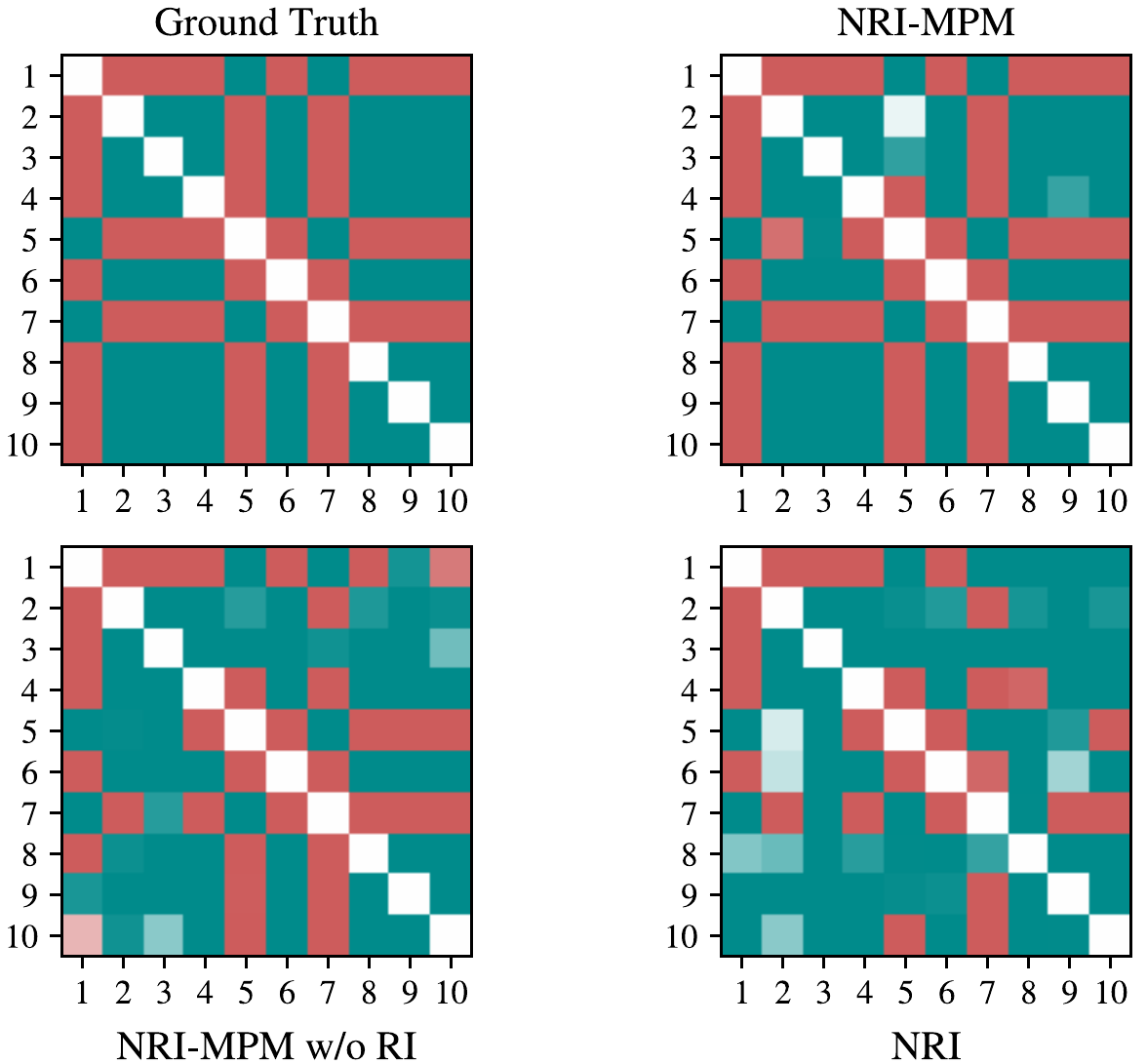}
  \caption{Predicted relations of different methods.}
  \label{fig:relation}
\end{figure}
\paragraph{Running Time of Different Methods}
The running times of different methods in a single epoch are reported in Fig.~\ref{fig:runtime}. It can be seen that our method requires some more time than NRI, which is a natural consequence of more complex models. {The running time of SUGAR is comparable with our method.} It is worth noting that ModularMeta requires much more running time than the others and the situation becomes more severe in larger systems. Maybe meta-learning the proposal function is computationally costly in ModularMeta. These results show that our method can achieve better performance with a smaller additional cost of computation.

\subsection{Ablation Study}
\begin{figure}[!t]
  \centering
  \includegraphics[width=0.85\columnwidth]{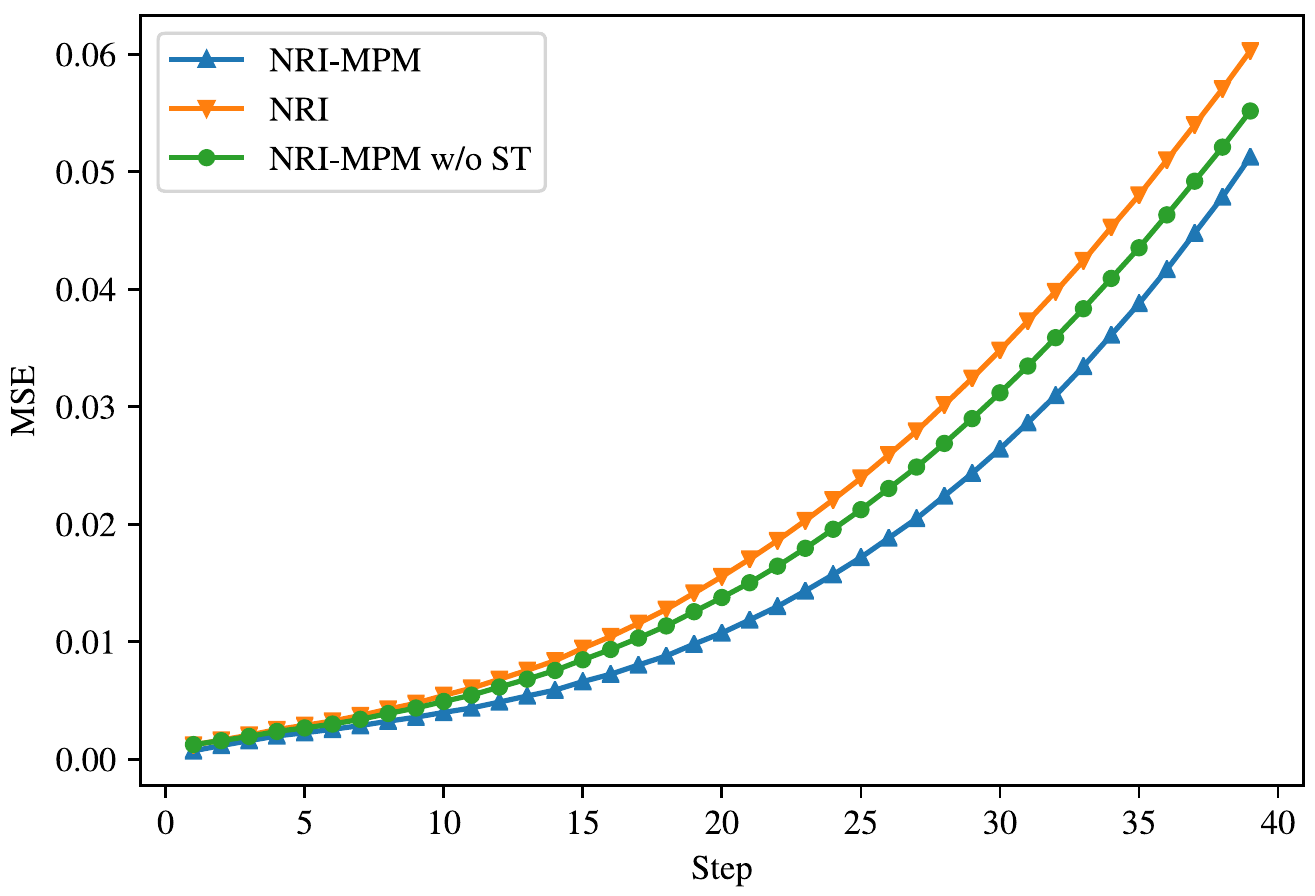}
  \caption{MSEs of multi-step prediction.}
  \label{fig:multistep}
\end{figure}
The ablation studies are conducted in the 10-object Charged dataset, and the results are shown in Table~\ref{tab:ablation}.
\begin{table}[!t]
  \centering
  \caption{Ablation study in the 10-object Charged dataset.}
  \label{tab:ablation}%
  \resizebox{\columnwidth}{!}{
    \begin{tabular}{l|cccc}
    \toprule
        Metrics  & Accuracy  & \multicolumn{3}{c}{MSEs} \\
    \midrule
    Prediction steps &       & \multicolumn{1}{c}{1} & \multicolumn{1}{c}{10} & \multicolumn{1}{c}{20} \\
    \midrule
    NRI-MPM w/o RI & $77.5\pm 0.2$     & 8.07e-4     & 4.13e-3     & 1.13e-2 \\
    NRI-MPM w/o intra-RI & $80.5\pm 0.4$     & 7.83e-4     & 3.90e-3     & 1.26e-2 \\
    NRI-MPM w/o inter-RI & $78.3\pm 0.7$     & 7.71e-4     & 4.11e-3     & 1.13e-2 \\
    NRI-MPM w/o ST & $74.0\pm 0.5$     & 1.26e-3     & 4.92e-3     & 1.38e-2 \\
    NRI-MPM w/o Sym & $79.3\pm 0.7$     & 7.60e-4     & $\textbf{3.77e-3}$     & 1.06e-2 \\
    NRI-MPM w/ hard Sym & $80.4\pm 0.5$     & 7.96e-4     & 3.83e-3     & 1.04e-2 \\
    \midrule
    NRI-MPM  & $\textbf{81.6}\pm 0.2$     & $\textbf{7.52e-4}$     & 3.80e-3     & $\textbf{1.03e-2}$ \\
    \bottomrule
    \end{tabular}%
  }
\end{table}%
\paragraph{Effect of Relation Interaction Mechanism}
As shown in Table~\ref{tab:ablation}, the accuracy drops significantly by removing the relation interacting mechanism, verifying its effectiveness in relation prediction. Besides, the MSEs decrease, indicating that more accurate relation prediction can help with future state prediction. {Besides, the contribution of the inter-edge interaction is higher than that of the intra-edge interaction. Intuitively, the intra- and inter-edge operations capture local and global interactions, respectively, and maybe in this dataset, global interactions are more informative than local interactions.}

To gain an intuitive understanding of the effect of the relation interaction mechanism, we conduct a case study on the 10-object charged particle systems. The distributions of predicted relations of NRI-MPM, NRI-MPM w/o RI and NRI together with the ground truth relations are visualized in Fig.~\ref{fig:relation}. The two types of relations are highlighted in red and green, respectively, while the diagonal elements are all in white as there no self-loops. It is known that two particles with the same charge repel each other while those with the opposite charge attract each other. Consequently, for any particles $v_i, v_j$, and $v_k$, the relations $e_{ij}$ and $e_{ik}$ are correlated. As shown in Fig.~\ref{fig:relation}, our method can model the dependencies among all relations much better than NRI, while removing the relation interaction mechanism results in less consistent prediction.
\paragraph{Effect of Spatio-temporal Message Passing Mechanism}
As shown in Table~\ref{tab:ablation}, the MSEs increase by removing the spatio-temporal message passing mechanism, and the differences are more significant for $M=10$ and $M=20$, verifying that the spatio-temporal message passing mechanism can alleviate error accumulation in multi-step prediction. Furthermore, the MSEs of different methods that predict 40 steps into the future are shown in Fig.~\ref{fig:multistep}. Note that the differences among different methods narrow for large $M$. Maybe it is still challenging for all methods to handle error accumulation in the long run.

Interestingly, the decrease of NRI-MPM w/o ST in terms of accuracy is more significant than that of NRI-MPM w/o RI. Maybe the spatio-temporal message passing mechanism imposes strong dependencies of historical interactions, which indirectly helps with relation reconstruction.
\paragraph{Effect of Symmetric Prior}
\begin{figure}[!t]
  \centering
  \includegraphics[width=0.9\columnwidth]{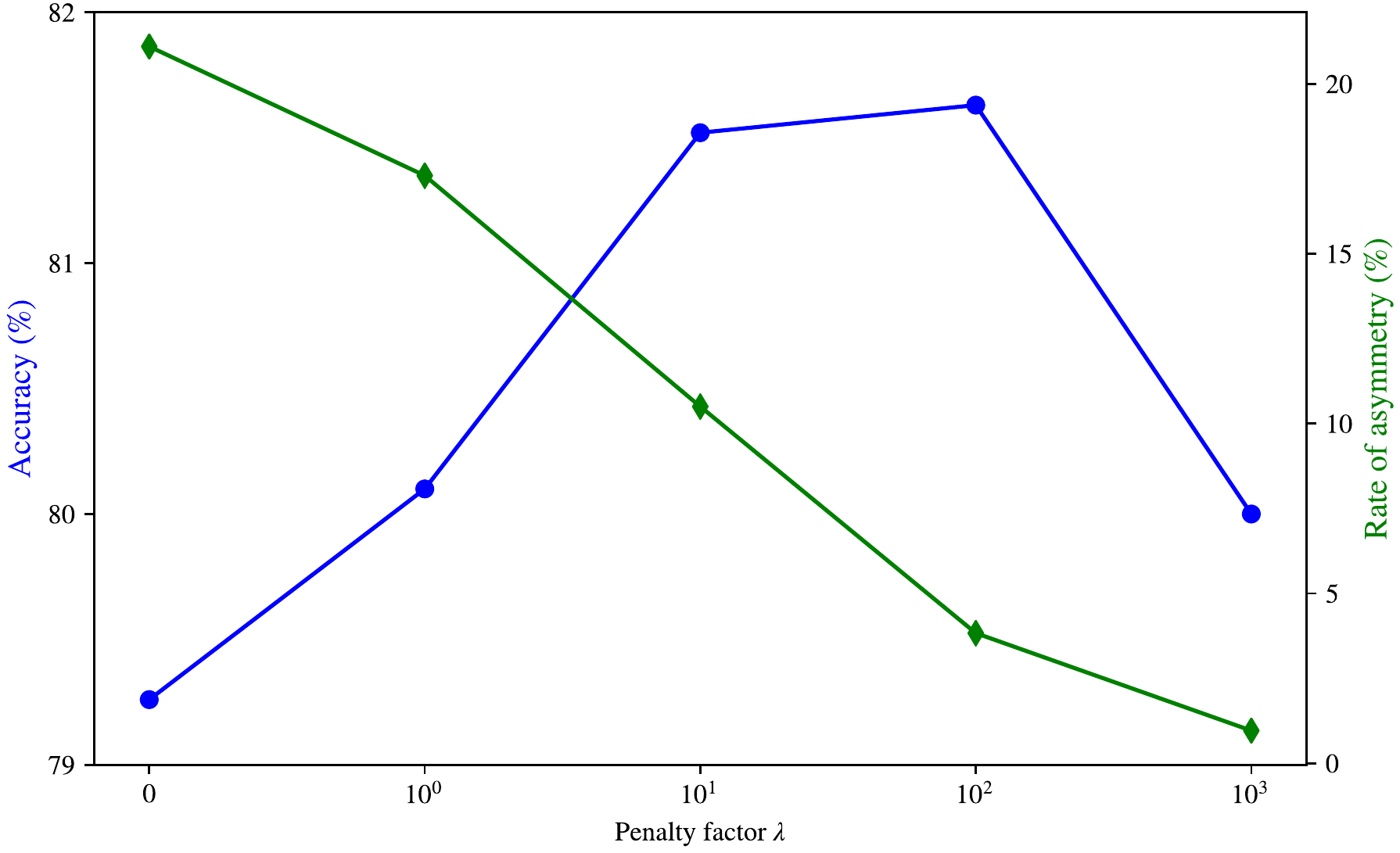}
  \caption{The accuracy and the rate of asymmetry w.r.t. $\lambda$.}
  \label{fig:lambda}
\end{figure}
As shown in Table~\ref{tab:ablation}, without the symmetric prior, the accuracy of NRI-MPM decreases by 2.3\%, while the MSEs are on par with the original model, indicating that the symmetric prior can help with relation reconstruction without hurting the precision of future state prediction. Compared with NRI-MPM w/ hard Sym, our method achieves higher accuracy with lower MSEs. Maybe the hard constraint of symmetry limits the exploration of the model in the training procedure, while a soft constraint provides more flexibility.

The effect of the penalty factor $\lambda$ is shown in Fig.~\ref{fig:lambda}. The rate of asymmetry decreases significantly as $\lambda$ increases, while the accuracy increases steadily and peaks around $\lambda=10^2$, verifying that adjusting $\lambda$ can control the effect of the symmetric prior and reasonable values will benefit relation reconstruction.

\section{Related Work}
This paper is part of an emerging direction of research attempting to model the explicit interactions in dynamical systems using neural networks as in NRI \cite{NRI}. 

Most closely related are the papers of Alet et al.~(\citeyear{ModularMeta}), Li et al.~(\citeyear{SUGAR}), Webb et al.~(\citeyear{fNRI}) and Zhang et al.~(\citeyear{GGN}). Alet et al.~(\citeyear{ModularMeta}) frame this problem as a modular meta-learning problem to jointly infer the relations and use the data more effectively. To deal with more complex systems, Li et al.~(\citeyear{SUGAR}) incorporate various structural prior knowledge as a complement to the observed states of agents. Webb et al.~(\citeyear{fNRI}) extend NRI to multiplex interaction graphs. Zhang et al.~(\citeyear{GGN}) explore relational inference over a wider class of dynamical systems, such as discrete systems and chaotic systems, assuming a shared interaction graph for all state sequences, which is different from the experimental settings of Kipf et al. (\citeyear{NRI}) and Alet et al. (\citeyear{ModularMeta}). Compared with this line of work, our method focuses on introducing efficient message passing mechanisms to enrich the representative power of NRI.

Many recent works seek to extend the message passing mechanisms of GNNs. Zhu et al.~(\citeyear{BiGNN}) define a bilinear aggregator to incorporate the possible interactions among all neighbors of a given node. Brockschmidt~(\citeyear{GNN-FiLM}) defines node aware transformations over messages to impose feature-wise modulation. Nevertheless, theses variants treat the messages as a transient part of the computation of the node embeddings, our relation interaction mechanism is aimed at learning edge embeddings. {Herzig et al.~(\citeyear{STICCV}) use non-local operations to capture the interactions among all relations, which requires quadratic time complexity.}

Besides, many works extend GNNs to handle structured time series. Graph convolutional networks with CNNs \cite{STGCN}, GRUs \cite{DCRNN}, or attention mechanisms \cite{ASTGCN} are introduced to deal with spatial and temporal dependencies separately for traffic flow forecasting. Furthermore, spatio-temporal graphs \cite{STSGCN} and spatio-temporal attention mechanisms \cite{GMAN} are proposed to capture complex spatio-temporal correlations. Our methods borrow these ideas from traffic flow forecasting to define spatio-temporal message passing operations for neural relational inference.

\section{Conclusion}
This paper introduces efficient message passing mechanisms with structural prior knowledge for neural relational inference. The relation interaction mechanism can effectively capture the coexistence of all relations and help make a joint prediction. By incorporating historical information, the spatio-temporal message passing mechanism can effectively alleviate error accumulation in multi-step state prediction. Additionally, the structural prior knowledge, symmetry as a special case, can promote the accuracy of relation reconstruction in more complex systems. The results of extensive experiments on simulated physics systems validate the effectiveness of our method. 

{Currently, only simple yet effective implementations using GRUs and attention mechanisms are adopted for the sequence models. Future work includes introducing more advanced models like the Transformers to further improve the performance. Besides, current experiments are conducted on simulated systems over static and homogeneous graphs. Future work includes extending our method to systems over dynamic and heterogeneous graphs.} Furthermore, the proposed method will be applied to study the mechanism of the emergence of intelligence in different complex systems, including multi-agent systems, swarm systems, physical systems and social systems.

\section{Ackownledgement}
This work is supported by the National Key R\&D Program of China (2018AAA0101203), and the National Natural Science Foundation of China (62072483, 61673403, U1611262). {This work is also supported by MindSpore.}

\bibliography{ref}

\begin{thebibliography}{33}
\providecommand{\natexlab}[1]{#1}
\providecommand{\url}[1]{\texttt{#1}}
\providecommand{\urlprefix}{URL }
\expandafter\ifx\csname urlstyle\endcsname\relax
  \providecommand{\doi}[1]{doi:\discretionary{}{}{}#1}\else
  \providecommand{\doi}{doi:\discretionary{}{}{}\begingroup
  \urlstyle{rm}\Url}\fi

\bibitem[{Alet et~al.(2019)Alet, Weng, Lozano-P{\'e}rez, and
  Kaelbling}]{ModularMeta}
Alet, F.; Weng, E.; Lozano-P{\'e}rez, T.; and Kaelbling, L.~P. 2019.
\newblock Neural Relational Inference with Fast Modular Meta-learning.
\newblock In \emph{NeurIPS}, 11827--11838.

\bibitem[{{Almaatouq} et~al.(2020){Almaatouq}, {Noriega-Campero}, {Alotaibi},
  {Krafft}, {Moussaid}, and {Pentland}}]{AdaptiveSocial}
{Almaatouq}, A.; {Noriega-Campero}, A.; {Alotaibi}, A.; {Krafft}, P.~M.;
  {Moussaid}, M.; and {Pentland}, A. 2020.
\newblock Adaptive Social Networks Promote the Wisdom of Crowds.
\newblock \emph{PNAS} 117(21): 11379--11386.

\bibitem[{{Bapst} et~al.(2020){Bapst}, {Keck}, {Grabska-Barwińska}, {Donner},
  {Cubuk}, {Schoenholz}, {Obika}, {Nelson}, {Back}, {Hassabis}, and
  {Kohli}}]{Glassy}
{Bapst}, V.; {Keck}, T.; {Grabska-Barwińska}, A.; {Donner}, C.; {Cubuk},
  E.~D.; {Schoenholz}, S.~S.; {Obika}, A.; {Nelson}, A. W.~R.; {Back}, T.;
  {Hassabis}, D.; and {Kohli}, P. 2020.
\newblock Unveiling the Predictive Power of Static Structure in Glassy Systems.
\newblock \emph{Nature Physics} 16(4): 448--454.

\bibitem[{Battaglia et~al.(2016)Battaglia, Pascanu, Lai, Rezende, and
  kavukcuoglu}]{Interaction-Networks}
Battaglia, P.; Pascanu, R.; Lai, M.; Rezende, D.~J.; and kavukcuoglu, K. 2016.
\newblock Interaction Networks for Learning about Objects, Relations and
  Physics.
\newblock In \emph{NeurIPS}, 4509–4517.

\bibitem[{{Brockschmidt}(2020)}]{GNN-FiLM}
{Brockschmidt}, M. 2020.
\newblock GNN-FiLM: Graph Neural Networks with Feature-wise Linear Modulation.
\newblock In \emph{ICML}.

\bibitem[{Cho et~al.(2014)Cho, Van~Merrienboer, Gulcehre, Bahdanau, Bougares,
  Schwenk, and Bengio}]{GRU}
Cho, K.; Van~Merrienboer, B.; Gulcehre, C.; Bahdanau, D.; Bougares, F.;
  Schwenk, H.; and Bengio, Y. 2014.
\newblock Learning Phrase Representations Using RNN Encoder--Decoder for
  Statistical Machine Translation.
\newblock In \emph{EMNLP}, 1724--1734.

\bibitem[{{Gilmer} et~al.(2017){Gilmer}, {Schoenholz}, {Riley}, {Vinyals}, and
  {Dahl}}]{MPNN}
{Gilmer}, J.; {Schoenholz}, S.~S.; {Riley}, P.~F.; {Vinyals}, O.; and {Dahl},
  G.~E. 2017.
\newblock Neural Message Passing for Quantum Chemistry.
\newblock In \emph{ICML}, 1263--1272.

\bibitem[{Guo et~al.(2019)Guo, Lin, Feng, Song, and Wan}]{ASTGCN}
Guo, S.; Lin, Y.; Feng, N.; Song, C.; and Wan, H. 2019.
\newblock Attention Based Spatial-Temporal Graph Convolutional Networks for
  Traffic Flow Forecasting.
\newblock In \emph{AAAI}, 922--929.

\bibitem[{Ha and Jeong(2020)}]{DataDivenComplex}
Ha, S.; and Jeong, H. 2020.
\newblock Towards Automated Statistical Physics: Data-Driven Modeling of
  Complex Systems with Deep Learning.
\newblock \emph{arXiv preprint arXiv:2001.02539} .

\bibitem[{Hamilton, Ying, and Leskovec(2017)}]{GraphSage}
Hamilton, W.; Ying, Z.; and Leskovec, J. 2017.
\newblock Inductive Representation Learning on Large Graphs.
\newblock In \emph{NeurIPS}, 1024--1034.

\bibitem[{{Herzig} et~al.(2019){Herzig}, {Levi}, {Xu}, {Gao}, {Brosh}, {Wang},
  {Globerson}, and {Darrell}}]{STICCV}
{Herzig}, R.; {Levi}, E.; {Xu}, H.; {Gao}, H.; {Brosh}, E.; {Wang}, X.;
  {Globerson}, A.; and {Darrell}, T. 2019.
\newblock Spatio-Temporal Action Graph Networks.
\newblock In \emph{ICCVW}.

\bibitem[{Hoshen(2017)}]{Vain}
Hoshen, Y. 2017.
\newblock VAIN: Attentional Multi-agent Predictive Modeling.
\newblock In \emph{NeurIPS}, 2701--2711.

\bibitem[{{Kingma} and {Welling}(2014)}]{VAE}
{Kingma}, D.~P.; and {Welling}, M. 2014.
\newblock Auto-Encoding Variational Bayes.
\newblock In \emph{ICLR}.

\bibitem[{{Kipf} et~al.(2018){Kipf}, {Fetaya}, {Wang}, {Welling}, and
  {Zemel}}]{NRI}
{Kipf}, T.; {Fetaya}, E.; {Wang}, K.-C.; {Welling}, M.; and {Zemel}, R. 2018.
\newblock Neural Relational Inference for Interacting Systems.
\newblock In \emph{ICML}, 2688--2697.

\bibitem[{Kuramoto(1975)}]{Kuramoto}
Kuramoto, Y. 1975.
\newblock Self-Entrainment of A Population of Coupled Non-Linear Oscillators.
\newblock In \emph{International Symposium on Mathematical Problems in
  Theoretical Physics}, 420--422. Springer.

\bibitem[{Lee et~al.(2019)Lee, Lee, Kim, Kosiorek, Choi, and
  Teh}]{SetTransformer}
Lee, J.; Lee, Y.; Kim, J.; Kosiorek, A.; Choi, S.; and Teh, Y.~W. 2019.
\newblock Set Transformer: A Framework for Attention-based
  Permutation-Invariant Neural Networks.
\newblock In \emph{ICML}, 3744--3753.

\bibitem[{Li et~al.(2020)Li, Yang, Tomizuka, and Choi}]{EvolveGraph}
Li, J.; Yang, F.; Tomizuka, M.; and Choi, C. 2020.
\newblock EvolveGraph: Multi-Agent Trajectory Prediction with Dynamic
  Relational Reasoning.
\newblock In \emph{NeurIPS}.

\bibitem[{Li et~al.(2019)Li, Meng, Shahabi, and Liu}]{SUGAR}
Li, Y.; Meng, C.; Shahabi, C.; and Liu, Y. 2019.
\newblock Structure-Informed Graph Auto-Encoder for Relational Inference and
  Simulation.
\newblock In \emph{ICML Workshop on Learning and Reasoning with
  Graph-Structured Representations}.

\bibitem[{{Li} et~al.(2018){Li}, {Yu}, {Shahabi}, and {Liu}}]{DCRNN}
{Li}, Y.; {Yu}, R.; {Shahabi}, C.; and {Liu}, Y. 2018.
\newblock Diffusion Convolutional Recurrent Neural Network: Data-Driven Traffic
  Forecasting.
\newblock In \emph{ICLR}.

\bibitem[{Maddison, Mnih, and Teh(2017)}]{Gumbel-softmax}
Maddison, C.~J.; Mnih, A.; and Teh, Y.~W. 2017.
\newblock The Concrete Distribution: A Continuous Relaxation of Discrete Random
  Variables.
\newblock In \emph{ICLR}.

\bibitem[{{Oliveira} et~al.(2020){Oliveira}, {Pinheiro}, {Macedo},
  {Bastos-Filho}, and {Menezes}}]{SwarmIntel}
{Oliveira}, M.; {Pinheiro}, D.; {Macedo}, M.; {Bastos-Filho}, C.; and
  {Menezes}, R. 2020.
\newblock Uncovering the Social Interaction Network in Swarm Intelligence
  Algorithms.
\newblock \emph{Applied Network Science} 5(1): 1--20.

\bibitem[{Sanchez-Gonzalez et~al.(2018)Sanchez-Gonzalez, Heess, Springenberg,
  Merel, Riedmiller, Hadsell, and Battaglia}]{GNs-block}
Sanchez-Gonzalez, A.; Heess, N.; Springenberg, J.~T.; Merel, J.; Riedmiller,
  M.; Hadsell, R.; and Battaglia, P. 2018.
\newblock Graph Networks as Learnable Physics Engines for Inference and
  Control.
\newblock In \emph{ICML}, 4467--4476.

\bibitem[{{Song} et~al.(2020){Song}, {Lin}, {Guo}, and {Wan}}]{STSGCN}
{Song}, C.; {Lin}, Y.; {Guo}, S.; and {Wan}, H. 2020.
\newblock Spatial-Temporal Synchronous Graph Convolutional Networks: A New
  Framework for Spatial-Temporal Network Data Forecasting.
\newblock \emph{AAAI} 914--921.

\bibitem[{van Steenkiste et~al.(2018)van Steenkiste, Chang, Greff, and
  Schmidhuber}]{RelationalEM}
van Steenkiste, S.; Chang, M.; Greff, K.; and Schmidhuber, J. 2018.
\newblock Relational Neural Expectation Maximization: Unsupervised Discovery of
  Objects and their Interactions.
\newblock In \emph{ICLR}.

\bibitem[{Watters et~al.(2017)Watters, Zoran, Weber, Battaglia, Pascanu, and
  Tacchetti}]{VisualNet}
Watters, N.; Zoran, D.; Weber, T.; Battaglia, P.; Pascanu, R.; and Tacchetti,
  A. 2017.
\newblock Visual Interaction Networks: Learning A Physics Simulator from Video.
\newblock In \emph{NeurIPS}, 4539--4547.

\bibitem[{Webb et~al.(2019)Webb, Day, Andres-Terre, and Li{\'o}}]{fNRI}
Webb, E.; Day, B.; Andres-Terre, H.; and Li{\'o}, P. 2019.
\newblock Factorised Neural Relational Inference for Multi-Interaction Systems.
\newblock \emph{ICML Workshop on Learning and Reasoning with Graph-Structured
  Representations} .

\bibitem[{{Xu} et~al.(2019){Xu}, {Hu}, {Leskovec}, and {Jegelka}}]{HowPowerful}
{Xu}, K.; {Hu}, W.; {Leskovec}, J.; and {Jegelka}, S. 2019.
\newblock How Powerful are Graph Neural Networks.
\newblock In \emph{ICLR}.

\bibitem[{{Yang} et~al.(2018){Yang}, {Yu}, {Bai}, {Wen}, {Zhang}, and
  {Wang}}]{Population}
{Yang}, Y.; {Yu}, L.; {Bai}, Y.; {Wen}, Y.; {Zhang}, W.; and {Wang}, J. 2018.
\newblock A Study of AI Population Dynamics with Million-agent Reinforcement
  Learning.
\newblock In \emph{AAMAS}, 2133--2135.

\bibitem[{{Yu}, {Yin}, and {Zhu}(2018)}]{STGCN}
{Yu}, B.; {Yin}, H.; and {Zhu}, Z. 2018.
\newblock Spatio-Temporal Graph Convolutional Networks: A Deep Learning
  Framework for Traffic Forecasting.
\newblock In \emph{IJCAI}, 3634--3640.

\bibitem[{Zhang et~al.(2020)Zhang, Wang, Xia, Lin, and Tong}]{SocialSystem}
Zhang, J.; Wang, W.; Xia, F.; Lin, Y.-R.; and Tong, H. 2020.
\newblock Data-Driven Computational Social Science: A Survey.
\newblock \emph{Big Data Research} 100145.

\bibitem[{{Zhang} et~al.(2019){Zhang}, {Zhao}, {Liu}, {Wang}, {Tao}, {Xin}, and
  {Zhang}}]{GGN}
{Zhang}, Z.; {Zhao}, Y.; {Liu}, J.; {Wang}, S.; {Tao}, R.; {Xin}, R.; and
  {Zhang}, J. 2019.
\newblock A General Deep Learning Framework for Network Reconstruction and
  Dynamics Learning.
\newblock \emph{Applied Network Science} 4(1): 1--17.

\bibitem[{{Zheng} et~al.(2020){Zheng}, {Fan}, {Wang}, and {Qi}}]{GMAN}
{Zheng}, C.; {Fan}, X.; {Wang}, C.; and {Qi}, J. 2020.
\newblock GMAN: A Graph Multi-Attention Network for Traffic Prediction.
\newblock In \emph{AAAI}, 1234--1241.

\bibitem[{Zhu et~al.(2020)Zhu, Feng, He, Wang, Li, Zheng, and Zhang}]{BiGNN}
Zhu, H.; Feng, F.; He, X.; Wang, X.; Li, Y.; Zheng, K.; and Zhang, Y. 2020.
\newblock Bilinear Graph Neural Network with Neighbor Interactions.
\newblock In \emph{IJCAI}, 1452--1458.

\end{thebibliography}


\section*{Supplementary material}

# Appendix A

Table A1: Notations Used in GNNs.

| Notations | Descriptions |
|---|---|
| $\mathcal{G} = (\mathcal{V}, \mathcal{E})$ | A graph $\mathcal{G}$ with a set of nodes $\mathcal{V}$ and a set of edges $\mathcal{E}$. |
| $v_j, e_{ij}$ | A node $v_j \in \mathcal{V}$ and a directed edge $e_{ij} \in \mathcal{E}$ from $v_i$ to $v_j$. |
| $\mathcal{N}_j$ | The set of indices of neighboring nodes with an incoming edge connected to $v_j$. |
| $v \to e, e \to v$ | The node-to-edge and edge-to-node message operations, respectively. |
| $f_v, f_e$ | Node-specific and edge-specific neural networks, respectively. |
| $\mathbf{x}_j, \mathbf{r}_{ij}$ | The features of node $v_j$ and edge $e_{ij}$, respectively. |
| $\mathbf{h}_i^l, \mathbf{h}_{ij}^l$ | The embeddings of node $v_i$ and edge $e_{ij}$ in the $l$-th layer of the GNNs, respectively. |

Table A2: Notations Used in NRI.

| Notations | Descriptions |
|---|---|
| $\mathbf{x} = (\mathbf{x}_1^{1:T}, ..., \mathbf{x}_N^{1:T})$ | The observed trajectories of $N$ objects in the system in $T$ time steps. |
| $\mathbf{x}_j^t, \Delta\mathbf{x}_j^t, \boldsymbol{\mu}_j^t$ | The state of node $v_j$, its change in state, and the predicted future state at time step $t$, respectively. |
| $\mathbf{z}$ | The latent relation matrix. |
| $\mathbf{z}_{ij}, z_{ij,k}$ | $\mathbf{z}_{ij}$ is a vector representing the edge type of $e_{ij}$ with $z_{ij,k}$ as its $k$-th elements. |
| $\hat{f}_e^k$ | A neural network modeling the effect of edge type $k$. |
| $\hat{f}_v$ | A neural network predicting the change in state $\Delta\mathbf{x}_j^t$. |
| $q_\phi, p_\theta$ | The encoder and the decoder, respectively. |
| $f_{\text{enc}}, f_{\text{dec}}$ | Synonyms for the encoder and the decoder, respectively. |
| $M$ | The number of time steps to predict. |

Table A3: Notations Used in NRI-MPM.

| Notations | Descriptions |
|---|---|
| $g_e, g_e^{\text{intra}}, g_e^{\text{inter}}$ | The relation interaction operation, intra-edge and inter-edge interaction operations, respectively. |
| $e \to e, v \to v$ | The edge-to-edge and node-to-node operations that directly pass messages among edges and nodes, respectively. |
| $\mathcal{I}_j, \mathcal{I}_j'$ | The ascending sequence of $\mathcal{N}_j$ and one of its permuted sequences, respectively. |
| $S, S_{\text{edge}}, S_{\text{node}}$ | Sequence models used in the intra-edge interaction operation, the node-to-edge and edge-to-node spatio-temporal message passing operations, respectively. |
| $\mathcal{S}_j$ | The set of all permutations on $\mathcal{I}_j$. |
| $\pi, \pi^{-1}$ | A permutation and its inverse, respectively. |
| $\mathbf{e}_{(i,j)}^1, \mathbf{e}_j^2$ | The result of $g_e^{\text{intra}}$ for edge $e_{ij}$ and that of $g_e^{\text{inter}}$ for node $v_j$, respectively. |
| $\mathbf{e}_j^1$ | The mean value of $\left\{\mathbf{e}_{(i,j)}^1\right\}_{i \in \mathcal{N}_j}$. |
| $\mathbf{e}_{(i,j)}$ | The output of the relation interaction mechanism for edge $e_{ij}$. |
| $\hat{\mathbf{e}}_{(i,j)}^t$ | The edge embedding that reflect the potential interaction between nodes $v_i$ and $v_j$ at time step $t$. |
| $\hat{\mathbf{h}}_{(i,j)}^t$ | The output of SeqModel$_{\text{edge}}$ for edge $e_{ij}$ at time step $t$. |
| $\text{MSG}_j^t$ | The aggregated spatial dependencies for node $v_j$ at time step $t$. |
| $\hat{\mathbf{h}}_j^t$ | The output of SeqModel$_{\text{node}}$ for node $v_j$ at time step $t$. |
| $q_\phi'$ | An auxiliary distribution of edge types such that $q_\phi'(\mathbf{z}_{ij}|\mathbf{x}) = q_\phi(\mathbf{z}_{ji}|\mathbf{x})$. |
| $\lambda$ | The penalty factor for the symmetric prior. |

## Appendix B

### Computation of the Loss Function

The loss function in Eq. (31) contains three terms, the reconstruction error $\mathbb{E}_{q_\phi(\mathbf{z}|\mathbf{x})}[\log p_\theta(\mathbf{x}|\mathbf{z})]$, KL divergence $\mathrm{KL}[q_\phi(\mathbf{z}|\mathbf{x})||p_\theta(\mathbf{z})]$ and the regularization term for symmetric prior $\mathrm{KL}\left[q'_\phi(\mathbf{z}|\mathbf{x})||q_\phi(\mathbf{z}|\mathbf{x})\right]$. As in NRI, the reconstruction error is estimated by

$$-\sum_j \sum_{t=2}^T \frac{\left\|\mathbf{x}_j^t - \boldsymbol{\mu}_j^t\right\|_2^2}{2\sigma^2} + \text{const}, \tag{A1}$$

and the KL divergence for a uniform prior is computed by

$$-\sum_{i \neq j} \sum_{k=1}^K q_\phi(z_{ij,k} = 1|\mathbf{x}) \log q_\phi(z_{ij,k} = 1|\mathbf{x}) + \text{const}. \tag{A2}$$

The regularization term is computed by

$$\sum_{i \neq j} \sum_{k=1}^K q'_\phi(z_{ij,k} = 1|\mathbf{x}) \log \frac{q'_\phi(z_{ij,k} = 1|\mathbf{x})}{q_\phi(z_{ij,k} = 1|\mathbf{x})}. \tag{A3}$$

## Appendix C

### Pseudo Code of NRI-MPM

---
**Algorithm 1:** NRI-MPM

---
**Input:** encoder $q_\phi$, decoder $p_\theta$, observed state sequences $\mathbf{x}^{1:T}$, time steps $M$, penalty factor $\lambda$.
**Output:** encoder $q_\phi$, decoder $p_\theta$.
1 Initialize $q_\phi$ and $p_\theta$ with random weights $\phi$ and $\theta$, respectively;
2 **repeat**
3     Compute the inferred relations $\mathbf{z}$ based on Eqs. (8)-(18);
    /* Split the observed state sequences into $D$ segments with length $M$. */
4     Compute $D = \lceil T/M \rceil$;
5     **for** $d = 1 : D$ **do**
6         **for** $m = 1 : M$ **do**
            /* Perform $M$-step prediction in each segment $d$. */
7             Compute $\boldsymbol{\mu}^{(d-1) \times M + m + 1}$ based on Eqs. (21)-(27);
8         **end**
9     **end**
10     Compute the auxiliary distribution $q'_\phi(\mathbf{z}|\mathbf{x})$ based on Eq. (28);
11     Compute the loss function $\mathcal{L}'$ based on Eq. (29);
12     Update $\phi$ and $\theta$ via gradient descent;
13 **until** *convergence or reach maximum number of iterations*;
14 **return** *encoder $q_\phi$, decoder $p_\theta$.*

---

## Appendix D

### Implementations of Relation Interaction Mechanisms

The sequence models of $g^{\text{intra}}$ in Eq. (14) and $g^{\text{inter}}$ in Eq. (16) are all implemented by GRUs. Let $\pi(\mathcal{I}_j) = \{i'_k\}_{k=1}^{|\mathcal{N}_j|}$ be a permuted sequence of $\mathcal{I}_j$, then SeqModel of $g^{\text{intra}}$ is defined as

$$\mathbf{e}^1_{(i'_1, j)} = \mathrm{GRU}\left(\mathbf{h}^2_{(i'_1, j)}, \mathbf{0}\right), \tag{A4}$$

$$\mathbf{e}^1_{(i'_k, j)} = \mathrm{GRU}\left(\mathbf{h}^2_{(i'_k, j)}, \mathbf{e}^1_{(i'_{k-1}, j)}\right), \quad k = 2, \ldots, |\mathcal{N}_j| \tag{A5}$$

where $\mathbf{0}$ is a zero vector.

Similarly, let $\{j_k\}_{k=1}^N$ be a permuted sequence of $\{j\}_{j=1}^N$, then the sequence model of $g^{\text{inter}}$ is defined as

$$\mathbf{e}_{j_1}^2 = \text{GRU}\left(\mathbf{e}_{j_1}^1, \mathbf{0}\right), \tag{A6}$$

$$\mathbf{e}_{j_k}^2 = \text{GRU}\left(\mathbf{e}_{j_k}^1, \mathbf{e}_{j_{k-1}}^2\right), \quad k = 2, \ldots, N. \tag{A7}$$

**Implementations of Spatio-temporal Message Passing Mechanisms**

This paper considers two simple implementations of the spatio-temporal message passing mechanisms, GRUs based implementations and attention based implementations. Both implementations consider $M$-step future state prediction with an initial time step $t_0$. One can combine the two implementations to obtain more complex mechanisms.

**GRUs Based Implementations** Similar to the implementations of sequence models in the relation interaction mechanism, $S_{\text{edge}}$ in Eq. (22) can be implemented by GRUs, namely,

$$\hat{\mathbf{h}}_{(i,j)}^{t_0+1} = \text{MLP}\left(\hat{\mathbf{e}}_{(i,j)}^{t_0}\right), \tag{A8}$$

$$\hat{\mathbf{h}}_{(i,j)}^{t_0+m} = \text{GRU}\left(\hat{\mathbf{e}}_{(i,j)}^{t_0+m}, \hat{\mathbf{h}}_{(i,j)}^{t_0+m-1}\right), \quad m = 2, \ldots, M \tag{A9}$$

where $t_0$ is the initial time step, $M$ is the number of time steps to predict, and MLP is used to obtain the hidden state of $\hat{\mathbf{e}}_{(i,j)}^{t_0+1}$.

Similarly, one can define $S_{\text{node}}$ as

$$\hat{\mathbf{h}}_j^{t_0+1} = \text{MLP}\left([\text{MSG}_j^{t_0}, \mathbf{x}_j^{t_0}]\right), \tag{A10}$$

$$\hat{\mathbf{h}}_j^{t_0+m} = \text{GRU}\left([\text{MSG}_j^{t_0+m}, \mathbf{x}_j^{t_0+m}], \hat{\mathbf{h}}_j^{t_0+m-1}\right), \quad m = 2, \ldots, M. \tag{A11}$$

**Attention Based Implementations** The sequence models can also be implemented through attention mechanisms. SeqModel$_{\text{edge}}$ in Eq. (22) can be defined as

$$\hat{\mathbf{h}}_{(i,j)}^{t_0+1} = \text{MLP}_1\left(\hat{\mathbf{e}}_{(i,j)}^{t_0}\right), \tag{A12}$$

$$\hat{\mathbf{h}}_{(i,j)}^{t_0+m} = \sum_{k=1}^m \boldsymbol{\beta}_{t_0+m, t_0+k} \cdot \begin{cases} \text{MLP}_2\left(\hat{\mathbf{h}}_{(i,j)}^{t_0+k}\right), & k < m \\ \text{MLP}_2\left(\hat{\mathbf{e}}_{(i,j)}^{t_0+m}\right), & k = m \end{cases} \quad m = 2, \ldots, M \tag{A13}$$

where $\boldsymbol{\beta}_{t_0+m, t_0+k}$ is the normalized temporal attention score between time steps $t_0 + k$ and $t_0 + m$, indicating the importance of time step $t_0 + k$ to $t_0 + m$. Specifically, the temporal attention mechanism is defined as follows,

$$\boldsymbol{\alpha}_{t_0+m, t_0+n} = \frac{1}{\sqrt{d}} \left\langle \text{MLP}_3\left(\hat{\mathbf{e}}_{(i,j)}^{t_0+m}\right), \text{MLP}_4\left(\hat{\mathbf{h}}_{(i,j)}^{t_0+n}\right) \right\rangle, \tag{A14}$$

$$\boldsymbol{\beta}_{t_0+m, t_0+k} = \frac{\exp(\boldsymbol{\alpha}_{t_0+m, t_0+k})}{\sum_{n=1}^m \exp(\boldsymbol{\alpha}_{t_0+m, t_0+n})}, \quad m = 2, \ldots, M \tag{A15}$$

where $\boldsymbol{\alpha}_{t_0+m, t_0+k}$ is the unnormalized temporal attention score between time steps $t_0 + m$ and $t_0 + k$, $d$ is the dimension of $\text{MLP}_3\left(\hat{\mathbf{e}}_{(i,j)}^{t_0+m}\right)$, and $\langle \cdot, \cdot \rangle$ denotes the inner product operator.

Similarly, SeqModel$_{\text{node}}$ in Eq. (24) can be defined as

$$\hat{\mathbf{h}}_j^{t_0+1} = \text{MLP}_5\left([\text{MSG}_j^{t_0}, \mathbf{x}_j^{t_0}]\right), \tag{A16}$$

$$\mathbf{m}_j^{t_0+m} = \left[\text{MSG}_j^{t_0+m}, \mathbf{x}_j^{t_0+m}\right], \tag{A17}$$

$$\hat{\mathbf{h}}_j^{t_0+m} = \sum_{k=1}^m \boldsymbol{\beta}_{t_0+m, t_0+k} \cdot \begin{cases} \text{MLP}_6\left(\hat{\mathbf{h}}_j^{t_0+k}\right), & k < m \\ \text{MLP}_6\left(\mathbf{m}_j^{t_0+k}\right), & k = m \end{cases} \quad m = 2, \ldots, M \tag{A18}$$

where

$$\boldsymbol{\alpha}_{t_0+m, t_0+n} = \frac{1}{\sqrt{d}} \left\langle \text{MLP}_7\left(\mathbf{m}_j^{t_0+m}\right), \text{MLP}_8\left(\hat{\mathbf{h}}_j^{t_0+n}\right) \right\rangle, \tag{A19}$$

$$\boldsymbol{\beta}_{t_0+m, t_0+k} = \frac{\exp(\boldsymbol{\alpha}_{t_0+m, t_0+k})}{\sum_{n=1}^m \exp(\boldsymbol{\alpha}_{t_0+m, t_0+n})}, \quad m = 2, \ldots, M. \tag{A20}$$

# Appendix E

**Experimental Settings**

**General Settings**  Unless otherwise noted, the settings are consistent with NRI.

- All methods are trained with a batch size of 128 for 500 epochs.
- The temperature $\tau$ in Eq. (3) is 0.5.
- The number of time steps $M$ to predict is 10 during the training procedure.
- The dimensions of all hidden layers of models are 256.
- All experiments are run 5 times on a machine containing 128GB of RAM, and 8 NVIDIA TITANXP graphics cards with PyTorch 1.2 and CUDA 10.0 in Ubuntu 16.04.

**Specific Settings for NRI-MPM**

- The initial learning rate is $2.5 \times 10^{-5}$ with a decaying factor 0.5 over every 200 epochs.
- The penalty factor $\lambda$ in Eq. (31) is selected from $\{1, 10, 10^2, 10^3\}$ via cross validation. In our experiments, $\lambda$ is set to $10^2$ except that it is set to 1 and $10^3$ for the 5-object and 10-object Kuramoto datasets, respectively.
- In all datasets, the relation interaction mechanism is implemented by GRUs as in Eqs. (A4)-(A7). The spatio-temporal message passing mechanism is implemented by GRUs as in Eqs. (A8)-(A11) except that in the Kuramoto dataset, it is implemented by combining GRUs and attention mechanisms as in Eqs. (A8)-(A20).

**MSEs in the 10-object Datasets**

Table A4: Mean squared error in predicting future states for simulations with 10 interacting objects.

| Datasets | Springs | | | Charged | | | Kuramoto | | |
|---|---|---|---|---|---|---|---|---|---|
| Predictions steps | 1 | 10 | 20 | 1 | 10 | 20 | 1 | 10 | 20 |
| LSTM | 3.46e-6* | 5.17e-4* | 3.21e-3* | 2.27e-3* | 1.00e-2* | 2.92e-2* | 6.08e-2* | 3.90e-1* | 4.24e-1* |
| NRI | 4.15e-7* | 3.74e-5* | 1.26e-4* | 1.20e-3* | 5.61e-3* | 1.65e-2* | 1.11e-2* | 7.80e-2* | 1.45e-1* |
| SUGAR | 4.68e-7* | 4.18e-5* | 1.37e-4* | 1.30e-3* | 5.17e-3* | 1.44e-2* | 1.25e-2* | 8.70e-2* | 1.61e-1* |
| ModularMeta | 3.60e-7* | 3.60e-5* | 1.12e-4* | 9.74e-4* | 5.85e-3* | 1.80e-2* | 1.48e-2* | 1.01e-1* | 1.71e-1* |
| NRI-MPM | **2.13e-7** | **1.41e-5** | **1.04e-4** | **7.52e-4** | **3.80e-3** | **1.03e-2** | **1.07e-2** | **5.40e-2** | **1.30e-1** |
| NRI-MPM (true graph) | 2.33e-9 | 1.03e-8 | 1.61e-7 | 8.06e-4 | 2.51e-3 | 5.66e-3 | 1.01e-2 | 1.58e-2 | 3.34e-2 |

* The results in these datasets are unavailable in the original paper, and they are obtained by running the codes provided by the authors.

The results of future state prediction in the 10-object datasets are shown in Table A4, and our method outperforms all baselines. Note that on average, the MSEs of all methods are larger than those in the 5-object datasets. Maybe learning the dynamics becomes more difficult in larger systems.


\end{document}